\documentclass[letterpaper,twocolumn,10pt]{article}
\usepackage{usenix}
\usepackage{balance}
\usepackage[skip=4pt]{caption}

\usepackage{titlesec}

\titlespacing*{\section}
  {0pt}{2.8ex plus 0.8ex minus 0.16ex}{1.84ex plus 0.16ex}

\titlespacing*{\subsection}
  {0pt}{2.6ex plus 0.8ex minus 0.16ex}{1.2ex plus 0.16ex}

\titlespacing*{\subsubsection}
  {0pt}{2.6ex plus 0.8ex minus 0.16ex}{1.2ex plus 0.16ex}

\usepackage{tikz}

\usepackage{microtype}

\usepackage{bm}

\usepackage{amsmath}

\usepackage{amssymb}

\usepackage{bbm}
\usepackage{graphicx}

\usepackage[ruled,noend]{algorithm2e}

\usepackage{color}

\usepackage{tcolorbox}
\usepackage{multirow}

\usepackage{makecell}
\usepackage{enumitem}
\usepackage{booktabs}
\usepackage{hyperref}
\usepackage{appendix}
\usepackage{indentfirst}
\usepackage{placeins}

\SetKwInOut{Parameter}{Parameters}

\begin{document}

\date{}

\title{\Large \bf Topic-FlipRAG: Topic-Orientated Adversarial Opinion Manipulation Attacks to Retrieval-Augmented Generation Models}


\author{
Yuyang Gong\textsuperscript{1}\thanks{Email: \texttt{2498002636gyy@gmail.com}},
Zhuo Chen\textsuperscript{1},
Jiawei Liu\textsuperscript{1}\thanks{Corresponding author. Email: \texttt{laujames2017@whu.edu.cn}},
Miaokun Chen\textsuperscript{1},
Fengchang Yu\textsuperscript{1},\\
Wei Lu\textsuperscript{1},
XiaoFeng Wang\textsuperscript{2},
Xiaozhong Liu\textsuperscript{3} \\
\textsuperscript{1}Wuhan University,
\textsuperscript{2}Nanyang Technological University,
\textsuperscript{3}Worcester Polytechnic Institute
}

\maketitle

\newcommand{\jiawei}[1]{\textcolor{blue}{JIAWEI: #1}}
\newcommand{\yuyang}[1]{\textcolor{orange}{YUYANG: #1}}
\newcommand{\cz}[1]{\textcolor{green}{CZ: #1}}
\begin{abstract}
Retrieval-Augmented Generation (RAG) systems based on Large Language Models (LLMs) have become essential for tasks such as question answering and content generation. However, their increasing impact on public opinion and information dissemination has made them a critical focus for security research due to inherent vulnerabilities. 
Previous studies have predominantly addressed attacks targeting factual or single-query manipulations. In this paper, we address a more practical scenario: topic-oriented adversarial opinion manipulation attacks on RAG models, where LLMs are required to reason and synthesize multiple perspectives, rendering them particularly susceptible to systematic knowledge poisoning.
Specifically, we propose Topic-FlipRAG, a two-stage  manipulation attack pipeline that strategically crafts adversarial perturbations to influence opinions across related queries. This approach combines traditional adversarial ranking attack techniques and leverages the extensive internal relevant knowledge and reasoning capabilities of LLMs to execute semantic-level perturbations.
Experiments show that the proposed attacks effectively shift the opinion of the model's outputs on specific topics, significantly impacting users' information perception. Current mitigation methods cannot effectively defend against such attacks, highlighting the necessity for enhanced safeguards for RAG systems, and offering crucial insights for LLM security research.
\end{abstract}

\section{Introduction}

\begin{figure}[!t]
  \centering
  \includegraphics[width=0.47\textwidth]{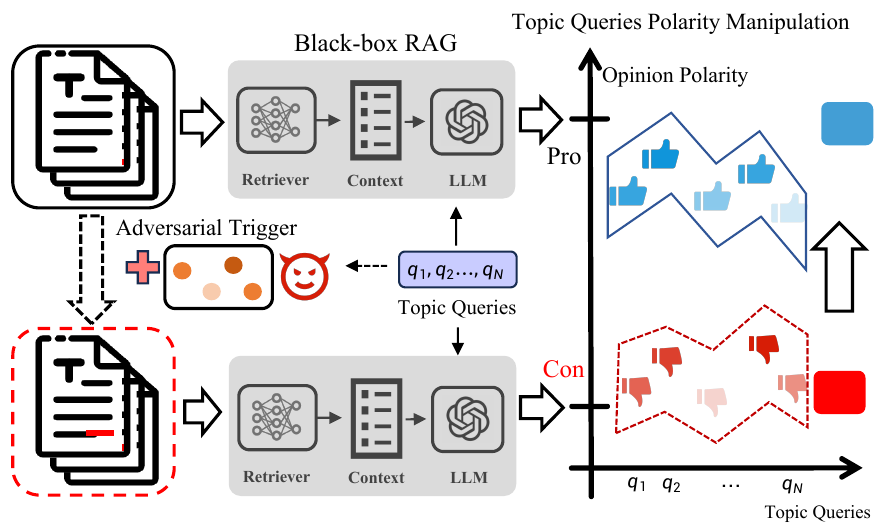}
  \caption{A concise overview of topic-oriented adversarial opinion manipulation attacks on RAG systems.}
  \label{intro:exam}
\end{figure}

Retrieval-Augmented Generation (RAG) systems, built on large language models (LLMs), have advanced significantly and been widely applied in tasks like question answering and content generation \cite{gao2023retrieval,zhao2024retrieval}. These systems integrate retrieval mechanisms with generative models, accessing diverse sources, e.g., Wikipedia, Reddit, and news articles, to deliver updated knowledge. While research has focused on RAG frameworks \cite{asai2023self,shi2025know,xu2024search}, the majority of studies emphasize performance and generalization. Consequently, security aspects have received comparatively little attention.

A defining feature of RAG systems is the vast scale of their referenced corpora or knowledge bases. However, these documents might originate from sources beyond service providers' control and may resist complete content purification. This creates new security risks for RAG-supported LLMs: adversaries can inject meticulously crafted malicious content \cite{carlini2024poisoning} into retrieval collections, ensuring its prioritized retrieval to influence LLM outputs. The inherent vulnerabilities of RAG models common to neural architectures, particularly their persistent impact on public opinion and information dissemination\cite{chen2025flipedrag,zhong2023poisoning,chen2024research}, have gradually drawn significant attention from the AI safety research community \cite{zhong2023poisoning,xue2024badrag,zou2024poisonedrag}. Investigating these vulnerabilities is critical for advancing RAG security and ensuring system robustness and credibility.

Early adversarial attacks on RAG systems primarily focused on jailbreak techniques and query-specific perturbations \cite{chen2025flipedrag,zhong2023poisoning}. These methods tamper with knowledge bases to manipulate retrieval rankings, enabling LLMs to provide answers defined by attackers for specific query-level, such as PoisonedRAG \cite{zou2024poisonedrag}, revealing critical vulnerabilities in applications like cybersecurity and healthcare. However, these attacks often lack practicality as they concentrate on isolated factual queries. Such attacks can often be countered through mitigation strategies such as reranking and filtering \cite{de2024rag,liu2023topic}. In contrast, thematic or topic-level attacks involve manipulating the overall perspective or stance of model outputs on broader queries, as shown in Figure \ref{intro:exam}. This form of manipulation is more practical and difficult to mitigate. Additionally, previous works overlook controversial topics, where limited user understanding may increase susceptibility to opinion manipulation.

This paper addresses this critical gap by concentrating on the nuanced manipulation of opinions through adversarial attacks targeted at topics within RAG models, which presents a novel and urgent challenge given the increasing complexity and reliance on these systems.
In line with settings from previous data poisoning findings\cite{zou2024poisonedrag,carlini2024poisoning}, we consider a scenario where the attacker could inject a few carefully crafted poisoned texts into the knowledge base. For example, if the knowledge base includes texts sourced from Wikipedia, the adversary might inject poisoned content by maliciously editing Wikipedia entries.
Besides, we focus on a more practical and challenging black-box scenario, where no model information is disclosed, except that the attackers can query the target RAG and obtain responses, which contains the corresponding candidate referential documents \cite{chen2025flipedrag}. 

In this paper, we propose Topic-FlipRAG, which is a two-stage multi-granularity attack method designed to manipulating the stance polarity of RAGs with camouflaged modification of target documents. Specifically, in the first stage, we perform a stealthy adversarial modifications on target documents. This is achieved by incorporating the extensive internal general semantic knowledge in LLMs with its analysis and reasoning ability.
In the second stage, inspired by adversarial semantic collision \cite{song2020adversarial}, we utilize gradients from an open-sourced neural ranking model (NRM) to generate topic-specific adversarial triggers.

Our experiments show Topic-FlipRAG achieves 0.5 average stance variation (ASV) across four domains, significantly outperforming other baselines. It demonstrates such adversarial attacks can effectively manipulate RAG models to produce outputs aligned with specific topical stances, thus influencing how information is presented and perceived. Based on that, we also conduct the user experiments, which reveal a significant impact on users’ opinion polarities toward controversial topics, with polarity shifts exceeding 16\% after interacting with poisoned RAG systems. Futhermore, we explore several potential mitigation strategies, including perplexity-based detection, random masking, paraphrasing and reranking. Our results show these mitigations are inadequate against adversarial opinion manipulation attack, especially for Topic-FlipRAG, underscoring the necessity for novel mitigation strategies.

Our major contributions are as follows:

(1) We explore a novel and practical security scenario, \textit{Topic-oriented RAG Opinion Manipulation}, which presents a broader and deeper threat to real-world users by enabling opinion manipulation across multiple topic-related queries.

(2) We propose Topic-FlipRAG, a knowledge-guided and multi-granularity black-box attack method tailored for topic-level opinion manipulation in RAG systems.

(3) Through extensive experiments, we demonstrate that Topic-FlipRAG significantly outperforms baseline methods across all evaluation metrics. Moreover, user studies reveal its substantial practical impact, highlighting its potential to manipulate user opinions.

(4) We analyze the effectiveness of several existing defense mechanisms and demonstrate their inadequacy in mitigating attacks by Topic-FlipRAG. These findings highlight the urgent need for more robust and adaptive defense strategies.

\section{Background and Related Work}

\subsection{Retrieval Augmented Generation (RAG)}
RAG models enhance their responses by accessing and incorporating external knowledge from large-scale databases or corpora during the generation process \cite{lewis2020retrieval,shi2025know,liu2023webglm}. By leveraging external data sources, RAG can provide more accurate and comprehensive answers, especially for queries requiring up-to-date information or specialized knowledge that may not be well-represented in the model's training data \cite{siyue2024mrag}. Moreover, it can scale more effectively and flexibly by updating the retrieval corpus without necessitating extensive retraining of the generative component \cite{gao2023retrieval,wu2024retrieval}.

The workflow of a RAG systems is generally divided into two sequential phases: \textit{retrieval} and \textit{generation}. 
In the retrieval phase, upon the submission of a user query $q$, the RAG system retrieves $k$ relevant documents from the corpus $D$ with the highest embedding similarities to the query $q$. Specifically, for each document $d \in D$, the relevance score with the query $q$ is computed as $R(q, d)$. In this paper, we adopt a more practical and widely used framework in LangChain
that the retrieval phase is incorporated with historical-aware query rewriting and intention reasoning. 
In the generation phase, Given a rewritten query $q'$ , a set of top-k retrieved documents $D_k$, and access to the LLM, one can query the LLM with $q'$ with the top-k retrieved documents $D_k$, the LLM generates an answer for original query $q$ by leveraging $D_k$ as context.

\subsection{Attacks to Retriever}
Retriever vulnerabilities in RAG systems pose significant security risks due to their neural ranking dependencies \cite{goren2018ranking}. Adversarial attacks manipulate document rankings through semantic perturbations to promote target documents in query-specific results \cite{liu2022order,wu2023prada}, undermining RAG's core assumption of reliable high-ranked contexts.
Current attacks are characterized along two axes: (1) Knowledge accessibility: White-box (full model access) vs. black-box (query-only) approaches. (2) Perturbation granularity: Word-level \cite{raval2020one,liu2023topic,wu2023prada}, Phrase-level \cite{song2020adversarial,liu2022order,chen2025flipedrag}, Sentence-level \cite{chen2023towards}, and Hybrid \cite{liu2024multi}.These neural semantic attacks parallel black-hat SEO tactics \cite{gyongyi2005web} but introduce novel challenges. Successful attacks propagate adversarial content to generators, enabling misinformation injection while evading traditional safeguards \cite{zhong2023poisoning,chen2025flipedrag}, necessitating integrated security frameworks for RAG systems.

\subsection{Attacks to LLMs}

Existing attacks on LLMs include jailbreak attacks \cite{wei2024jailbroken,deng2024masterkey,li2023multi,lin2024figure}, backdoor attacks \cite{kandpal2023backdoor,lu2024test,cao2023stealthy}, prompt injection \cite{liu2023prompt,perez2022ignore,greshake2023not}, and poisoning attacks \cite{shafran2024machine,zou2024poisonedrag,cho2024typos,zhong2023poisoning,chen2025flipedrag,zhang2024hijackrag}. Poisoning attacks uniquely threaten RAG systems by injecting adversarial content into retrieval corpora to manipulate outputs. Current RAG poisoning studies focus on closed-domain factoid QA pairs (e.g., "CEO of OpenAI") \cite{zou2024poisonedrag,zhong2023poisoning,cho2024typos,shafran2024machine,zhang2024hijackrag}, overlooking the multi-query nature of real-world topic exploration\cite{liu2023topic,xue2024badrag}, e.g., "smartwatch battery life" and "health tracking accuracy" under the "wearable tech" theme. This motivates our focus on practical topic-level universal perturbations mirroring universal adversarial examples.

Furthermore, existing defenses based on fact-checking are insufficient for opinion-based queries, such as "Should genetic testing be regulated?", which require nuanced reasoning rather than simple factual recall. Building on the insights from FlippedRAG \cite{chen2025flipedrag}, we tackle this significant gap by examining the adversarial manipulation of controversial topics. In such cases, LLMs are required to reason and synthesize multiple perspectives, making them particularly vulnerable to systematic knowledge poisoning.

\section{Threat Model}
\label{sec:threat_model}
Given a set of topic-queries $Q = \{q_1, q_2, \ldots, q_{|Q|}\}$, a RAG corpus $D$, and a desired target polarity $S_t$ (e.g., \textit{Pro} or \textit{Con}), The adversary's objective is to steer the RAG system's overall stance across the entire set of queries toward stance $S_t$. Let $\text{doc}_{\text{tar}}$ be the target document the attacker aims to promote. The black-box retrieval model $\text{RM}$ assigns a relevance score $R(q, d)$ to each document $d \in D$ and returns the top-$k$ documents as $\text{RM}_k(q)=\{d_1, d_2, \ldots, d_k\}$. The LLM then generates an answer $\text{LLM}(q, RM_k(q))$, and the extracted opinion from this response is $S_o = \text{S}(\text{LLM}(q, RM_k(q)))$.

\begin{figure*}[!t]
  \centering
  \includegraphics[width=0.89\textwidth]{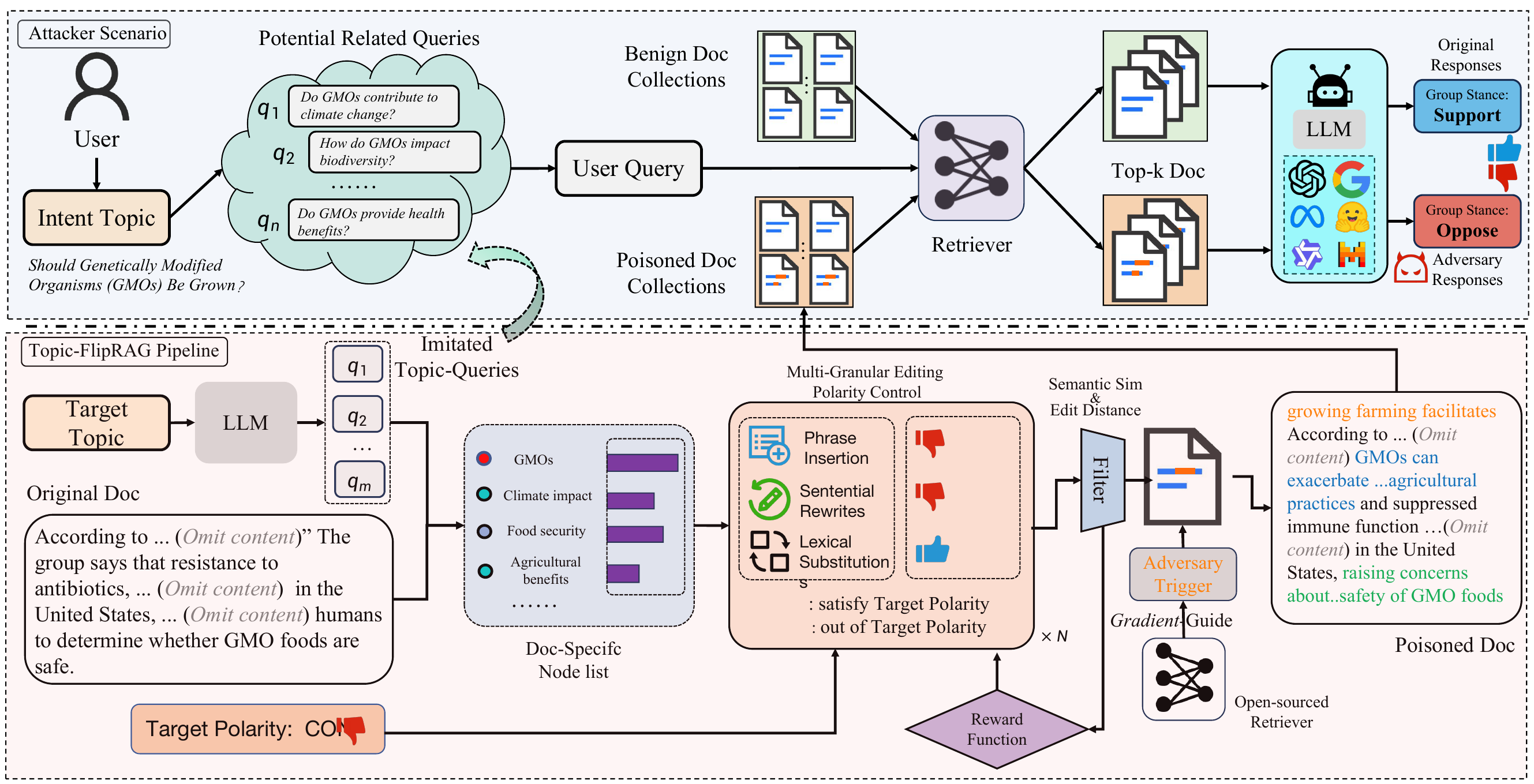}
  \caption{An overview of our proposed Topic-Orientated Adversarial Opinion Manipulation Attack method for RAG systems.}
  \label{overview}
\end{figure*}

\subsection{\textbf{Objective of the Adversary}}
\label{sec:obj_adversary}
The adversary seeks to modify the target document $doc_{tar}$ through a series of subtle perturbations $p_{adv}$, while preserving its original semantics and fluency, thereby transforming the corpus $D$ into $D' = D(\text{doc}; \text{doc}_{tar} \oplus p_{\text{adv}})$. By increasing the retrieval model’s relevance score $R(q, \text{doc}_{tar} \oplus p_{\text{adv}})$ for each query $q \in Q$, the promoted document is  more likely to appear among the top-$k$ results $\text{RM}_k(q)$.

As a result, when the LLM generates answers conditioned on the query and the top-$k$ retrieval results $(q, \text{RM}_k(q))$, the adversarially augmented document will influence the opinion $S(\text{LLM}(q, \text{RM}_k(q)))$ towards the target polarity $S_t$. Crucially, the objective extends beyond a single query: the adversary aims to ensure that, on average, the opinions produced by the LLM across the entire query set $Q$ align with $S_t$. In other words, the attacker seeks:
\begin{align*}
    \max_{p_{\text{adv}}} \frac{1}{|Q|}\sum_{q \in Q} I\bigl(S(\text{LLM}(q, \text{RM}_k(q))) = S_t\bigr) \hspace{5mm}
\end{align*}

where $I(\cdot)$ is an indicator function that returns $1$ if the generated stance matches $S_t$ and $0$ otherwise. By optimizing for this criterion, the adversary ensures that the aggregated stance across all topic queries shifts significantly toward the desired polarity.

\subsection{\textbf{Capabilities of the Adversary}}
In the black-box setting, the adversary is restricted to modifying a limited number of documents within the corpus $D$. Obtaining detailed information about the specific retrieval model employed by the target RAG system is typically impractical in real-world scenarios. The adversary has no access to the internal parameters or architecture of either the retrieval model or the large language model (LLM), and cannot alter the LLM’s prompt templates.
\section{Methodology}

\subsection{Overview}

In this study, we introduce \textbf{Topic-FlipRAG}, a two-stage, multi-granularity, topic-oriented attack pipeline designed to manipulate the stance polarity of RAG output, as illustrated in Figure~\ref{overview}. 
The pipeline proceeds as follows:

\textbf{Stage 1: Knowledge-Guided Attack.}
In the initial stage, we leverage the extensive internal knowledge and reasoning capabilities of LLMs to perform adversarial edits on the target document at a general semantic level. We identify and extract essential topic-related information nodes from the queries, then guide the LLM to integrate these nodes into \( \text{doc}_{tar} \) through a multi-granular editing strategy across three semantic dimensions. Simultaneously, rigorous polarity control is employed to preserve the original stance of the document, ensuring precise manipulation. A dynamic reward function guides each modification step, embedding critical elements and relevant information while minimizing disruption to the document’s original semantics. This process lays a solid foundation for the subsequent stage.

\textbf{Stage 2: Adversarial Trigger Generation.}
In the second stage, we enhance the similarity of the adversarially edited document \(\text{doc}_{know}\), which is generated in Stage 1. Leveraging an open-sourced Neural Ranking Model (NRM), we generate a concise adversarial trigger specifically crafted to maximize the alignment between \( \text{doc}_{know} \) and the topic queries. This trigger is then fused with \( \text{doc}_{know} \) and injected into the RAG system’s database as a poisoned document, substantially increasing its likelihood of being retrieved. Consequently, the generated output shifts toward the target stance polarity, thereby achieving the objective of topic-oriented adversarial manipulation.

\subsection{Knowledge-Guided Attack}
\label{sec:knowledge_guide_attack}

In this section, we propose a three-phase framework, termed the \textit{knowledge-guided attack} (know-attack), to systematically modify a target document \( \text{doc}_{tar} \) to enhance its relevance to a given query set. The framework seeks to minimally modify the original document while effectively integrating topic-specific nodes, thereby 
maximally influencing the final RAG output. By operating across multiple levels of modification, it ensures semantic coherence, adheres to minimal editing principles, and preserves a predefined stance, enabling precise and controlled manipulation of the target document.

\textbf{Problem Statement.}  
Given a target document $\text{doc}_{tar}$ and a topic-related query set $Q = \{q_1, q_2, \ldots, q_{|Q|}\}$, we seek a modified document $\text{doc}_{know} = \text{doc}_{tar} \oplus \mathcal{P}$ , where $\mathcal{P}$ denotes subtle perturbations, that increases its overall relevance to all queries while adhering to a target stance $S_t$.

\begin{equation*}
\max_{\mathcal{P}} \frac{1}{|Q|}\sum_{q \in Q} Sim(q,\text{doc}_{know})
\end{equation*} 

subject to:
\begin{equation*}
\|\mathcal{P}\| \leq \epsilon,\quad Sim(\text{doc}_{know}, \text{doc}_{tar}) \geq \lambda,\quad S(\text{doc}_{know}) = S_t.
\end{equation*}

Here, $\|\mathcal{P}\|$ constrains the degree of textual alteration to ensure minimal editing, and $Sim(\cdot)$ ensures sufficient semantic similarity between the original and modified documents. $\epsilon$ and $\lambda$ are predefined parameters used to constrain the extent of modification. By incorporating these constraints, the resulting \(\text{doc}_{know}\) not only aligns with the chosen stance $S_t$ and achieves higher relevance for all queries in the query set $Q$, but also maintains the document’s intrinsic coherence and readability, increasing its likelihood of influencing the RAG system’s outputs while remaining undetectable.

\subsubsection {\textbf{Phase 1: Key Node Extraction and Selection}}
Given a topic and its corresponding set of queries $Q = \{q_1, q_2, \ldots, q_{|Q|}\}$, we first leverage a LLM, e.g., GPT-4o-mini, to identify a set of $K$ key information nodes that are most salient for maximizing relevance with respect to the entire query set. These $K$ nodes form a candidate node list $\mathcal{L}$ encapsulating crucial topical aspects.

Next, we analyze $\text{doc}_{tar}$ to determine which nodes from $\mathcal{L}$ are underemphasized or entirely absent. The LLM generates a \emph{doc-specific node list} by providing explicit reasoning for each missing or underemphasized node. This doc-specific node list guides subsequent adversarial editing, ensuring that modifications focus on incorporating these critical nodes without introducing extraneous or tangential content.

\subsubsection{\textbf{Phase 2: Multi-Granular Adversarial Editing and Polarity Control}}
Inspired by \cite{liu2024multi} and adversarial strategies against black-box neural ranking models \cite{liu2022order,wu2023prada}, we adopt a multi-granularity editing approach that leverages the advanced language understanding capabilities of LLMs. The editing process operates on three levels: (1) \textbf{lexical substitutions}, replacing individual words with synonyms or node-relevant terms while preserving the original semantics; (2) \textbf{sentential rewrites}, restructuring sentences and injecting minimal node-related content without altering the intended meaning; and (3) \textbf{phrase insertions}, strategically adding brief, contextually appropriate sentences containing node information to maintain narrative coherence and content integrity.

During this adversarial editing process, we incorporate a module called \textbf{Polarity Control} to enforce a target stance \( S_t \) constraint, guiding the overall direction of the document. For instance, if the target polarity is set to CON (i.e., opposing a certain viewpoint), all modifications introduced by the model consistently reflect negative or critical perspectives. This module ensures that the integrated nodes and textual changes are aligned with the desired polarity, ultimately steering the RAG output toward the intended viewpoint.

\subsubsection{\textbf{Phase 3: Iteration with Rewarding Function and Final Output Selection}}

To adhere to the principle of minimal editing while ensuring effective document modification, we introduce an iterative framework guided by an \emph{augmentation factor} $t$. This factor dynamically controls the extent and frequency of adversarial modifications during each editing iteration. A larger $t$ encourages more extensive and aggressive edits, whereas a smaller $t$ leads to a more conservative editing strategy. To mitigate fluctuations caused by the stochasticity of LLMs, we perform $I$ repeated sampling runs for each value of $t$. It ensures stable performance by averaging across multiple candidate modifications. After each editing iteration, $t$ is adaptively adjusted based on feedback from a filtering mechanism to maintain a balance between relevance and minimal editing.

In each iteration $n$ ($n=1,\dots,N$), for a given $t_n$, we generate $I$ candidate modified documents $\{\text{doc}_{m}^{(n,i)}\}_{i=1}^I$. Each candidate document is evaluated against two critical metrics:

\textbf{Edit Distance (Edit Ratio)}: $d_{\mathrm{edit}}\bigl(\text{doc}_{m}, \text{doc}_{tar}\bigr)$ measures the extent of modification compared to the original document.

\textbf{Semantic Similarity}: $d_{\mathrm{sem}}\bigl(\text{doc}_{m}, \text{doc}_{tar}\bigr)$ evaluates how well the modified document preserves the original semantics.

To ensure minimal disruption, we define a strict edit ratio threshold $\epsilon$ and a semantic similarity threshold $\lambda$. A filtering indicator is then used to assess the validity of each candidate document based on these criteria:

{\small
\[
    \mathrm{Filter}\bigl(\text{doc}_{m}\bigr) 
    \;=\; 
    \begin{cases}
        1, & \begin{matrix}
        \text{if } (d_{\mathrm{edit}}\bigl(\text{doc}_{m}, \text{doc}_{tar}\bigr)\,\le\,\epsilon ) \\
        \;\;\wedge \;(d_{\mathrm{sem}}\bigl(\text{doc}_{m}, \text{doc}_{tar}\bigr)\,\le\,\lambda),
        \end{matrix} \\
        0, & \text{otherwise}.
    \end{cases}
\]
}

The augmentation factor $t$ is dynamically updated based on the feedback from the filtering layer. To balance the trade-off between under-editing and over-editing, we define a partial edit threshold $\rho = 0.75\,\epsilon$ and employ the update function:
{\small
\[
  t_{n+1} \;=\;
  \begin{cases}
    t_n - \delta, 
      & \text{if } \max\limits_{i} \mathrm{Filter}\bigl(\text{doc}_{m}^{(n,i)}\bigr) = 0, \\[6pt]
    t_n + \delta, 
      & \text{if } \exists\,
      \begin{matrix}
          ( i \text{ s.t. } \mathrm{Filter}\bigl(\text{doc}_{m}^{(n,i)}\bigr) = 1)\\ 
          \;\wedge\;
          (d_{\mathrm{edit}}\!\bigl(\text{doc}_{m}^{(n,i)}, \text{doc}_{tar}\bigr) < \rho), \end{matrix}
          \\
        t_n, 
      & \text{otherwise}.
  \end{cases}
\]}

Here, $\delta>0$ controls the aggressiveness of the adjustments. If no candidate satisfies the filtering constraints, $t$ is decreased to avoid excessive modifications. Conversely, if valid candidates exist but fail to meet the partial edit threshold $\rho$, $t$ is increased to promote more aggressive edits. Otherwise, $t$ remains unchanged to preserve stability.

After $N$ iterations, the process produces a set of $M$ valid candidate documents that satisfy the edit ratio and semantic similarity constraints. A final selection phase is then conducted to identify the optimal output. For each retained candidate $\text{doc}_m$, we use a NRM to estimate its relevance for each query $q_i \in Q$, denoted as $R\bigl(q_i, \text{doc}_m\bigr)$. The average relevance score for each candidate is computed as:
\[
\bar{R}\bigl(\text{doc}_m\bigr) \;=\; 
\frac{1}{|Q|}\,\sum_{i=1}^{|Q|} R\bigl(q_i, \text{doc}_m\bigr).
\]

The final modified document $\text{doc}_{know}$ is selected by maximizing the average relevance:
\[
\text{doc}_{know} 
\;=\;
\arg\max_{\text{doc}_m} \; \bar{R}\bigl(\text{doc}_m\bigr).
\]

The iterative process, directed by the rewarding function and filtering mechanisms, ensures that the final output $\text{doc}_{know}$ incorporates topic-specific nodes with minimal yet effective modifications. By maintaining semantic coherence and adhering to predefined constraints, it enhances the relevance of $\text{doc}_{know}$ to the query set while aligning its stance with $S_t$.

\subsection{Adversarial Gradient-Based Trigger Generation}
\label{sec:adv_trigger_generation}

Building on the know-attack introduced in Section~\ref{sec:knowledge_guide_attack}, which integrates general semantic-level node information into the target document \(\text{doc}_{tar}\), we further aim to generate adversarial triggers targeting the retrieval model within the RAG system. Specifically, we append a stealthy adversarial trigger \(T\) to the previously constructed \(\text{doc}_{know}\), resulting in the final adversarially augmented document \(\text{doc}_{adv}\). Inspired by adversarial semantic collision \cite{song2020adversarial,liu2022order}, we derive \(T\) through a gradient-based optimization process using a public neural ranking model\footnote{\url{https://huggingface.co/nboost/pt-bert-base-uncased-msmarco}}. By integrating \(\text{doc}_{know}\)  with \(T\), we align the document more closely with the query set $Q$, thereby increasing its likelihood of being retrieved. Once ranked among the top-K results, the strategically crafted content within \(\text{doc}_{adv}\) can effectively steer the LLM’s final output toward \(S_t\).

\subsubsection{Problem Formulation}

Our goal, as described in Section~\ref{sec:obj_adversary}, is to guide the RAG system toward consistently adopting the target polarity $S_t$ across the entire query set $Q$. However, optimizing this topic-level objective is non-trivial due to the absence of gradient signals from the LLM.

To tackle this challenge, we introduce an intermediate objective centered on document relevance. Let $RM(q_i, \text{doc})$ denote the relevance score assigned by the neural ranking model (NRM) for a query-document pair $(q_i, \text{doc})$. For a crafted document $\text{doc}_{adv} = [\text{doc}_{know}, T]$, our interim optimization goal is to find a trigger $T$ that maximizes the average relevance score across all queries in $Q$:
\[
\max_{T} \frac{1}{|Q|} \sum_{i=1}^{|Q|} RM\left(q_i, [\text{doc}_{know}, T]\right).
\]
By enhancing the document’s relevance to each $q \in Q$, we increase the likelihood that $\text{doc}_{adv}$ appears among the top-$k$ retrieved documents, thereby shaping the RAG system’s output polarity in favor of the target stance $S_t$.

\subsubsection{Gradient-Based Optimization}

We use a gradient-based search strategy to generate the trigger $T$ that optimizes the average relevance across the topic-related query set. The search proceeds in two steps: 1) we optimize a soft trigger in the continuous embedding space using gradients from the NRM, where the updates are performed using the Adam\cite{kingma2014adam} optimizer; and 2) we apply beam search to discretize the soft representation into a valid token sequence. These two steps are repeated iteratively until the relevance objective converges. Through this process, we obtain a discrete trigger that maximizes document relevance over all topic queries.

\textbf{Gradient Optimization:} During each optimization step, we compute and combine gradients for all queries in $Q$. Inspired by \cite{liu2022order}, we introduce a word enhancement mechanism that incorporates query-specific keywords and lexical distributions derived from the model’s vocabulary. This guided selection of candidate tokens captures semantic overlaps among multiple queries and enhances overall trigger relevance.

\textbf{Beam Search:} During the beam search phase, we evaluate candidate triggers, along with doc-specific augmentation, to select those that maximize combined similarity scores across the entire query set. This selection process ensures that the chosen trigger consistently enhances the ranking impact of the target passage when combined with \(\text{doc}_{know}\).

\textbf{Doc-Specific Augmentation:} At each iteration of gradient calculation and beam search, the intermediate soft representation of the trigger is evaluated only in the context of its concatenation with \(\text{doc}_{know}\) and jointly across all queries. While this introduces additional computational overhead, it ensures tight contextual alignment with the target document, leading to more substantial gains in retrieval performance.

\subsubsection{Final Output and RAG Opinion Influence}
After convergence, the adversarial trigger $T$ is appended to $\text{doc}_{know}$ to form the final adversarial document:
\[
\text{doc}_{adv} = [\text{doc}_{know}, T].
\]

This $\text{doc}_{adv}$ is then injected into the RAG system's corpus $D'$, effectively increasing its likelihood of being retrieved among the top-$k$ documents for each query $q \in Q$. Consequently, $\text{doc}_{adv}$ frequently appears in retrieval results, leading the LLM to generate outputs that align with the target stance $S_t$. This process effectively manipulates the overall stance of the RAG system's output by ensuring that the augmented document is both prominently retrieved and influences the generated responses toward the target stance $S_t$.

\section{Experiments}

\subsection{Research Questions}

We propose three research questions to evaluate the effectiveness of our method in the topic-queries task, focusing on black-box NRM attacks and opinion manipulation to RAGs.

\textbf{RQ1}: Can Topic-FlipRAG effectively boost the rankings of target documents across queries in the retriever within RAG?

\textbf{RQ2}: To what extent does Topic-FlipRAG affect the answers generated by the target RAG systems?

\textbf{RQ3}: Does topic-oriented opinion manipulation significantly impact users' perceptions of controversial topics?

\subsection{Datasets}

\textbf{MSMARCO}.
We utilized the MS MARCO Passage Ranking dataset \cite{nguyen2016ms} to evaluate the effectiveness of our method in improving document ranking in challenging topic-queries tasks. Specifically, we examined whether our method could effectively boost the retrieval rankings of target documents across queries in the retriever within RAG.

For topic-list construction, we applied K-means clustering to group similar queries, forming topics with related queries. To ensure robustness, we filtered topics based on intra-topic similarity, retaining only those with high semantic diversity and sufficient query count. This process resulted in 29 topics, each containing an average of 22.28 queries with a similarity score of approximately 0.5, thereby supporting a rigorous and diverse evaluation. 

\textbf{PROCON}.\label{procon}
We conducted our opinion manipulation experiments using a curated dataset of controversial topics sourced from the PROCON.ORG website. The dataset spans over 80 topics across diverse domains such as society, health, government, and education. Each topic is framed around two opposing stance labels \{\textit{PRO (support), CON (oppose)}\}, with accompanying passages that present arguments from both perspectives.

To simulate real-world user interactions with a RAG system, we instructed GPT-4o to act as a proxy user and generate 40 candidate sub-queries for each topic. These sub-queries were crafted to reflect the diverse questions and concerns that users might pose when engaging with a specific controversial topic. To ensure semantic diversity, we applied a similarity-based filtering process, retaining only sub-queries with pairwise similarity scores below approximately 0.85. This step effectively removed redundancies while preserving a broad spectrum of perspectives. As a result, the final set of topic queries achieved an average similarity score of approximately 0.7, balancing semantic relevance with diversity. More detailed information on the datasets refer to the dataset description in our github repository\footnote{\url{https://github.com/LauJames/Topic-FlipRAG}}.

\subsection{Experiment Details}
The specific setting details for the Topic-queries RAG manipulation experiment are as follows:

(1) Black-box RAG. The black-box RAG process is denoted as \( \text{RAG}_{\text{black}}\). The RAG framework is the Conversational RAG from LangChain for our experiments. The LLMs adopted in RAG are two widely adopted open-source instruction-tuned models: Meta-Llama-3.1-8B-Instruct\footnote{\url{https://huggingface.co/meta-llama/Llama-3.1-8B-Instruct}} (referred to as Llama3.1) and Qwen-2.5-7B-Instruct\footnote{\url{https://huggingface.co/Qwen/Qwen2.5-7B-Instruct}} (Qwen2.5).

(2) Retrieval model. We benchmark three dominant dense retrievers, i.e., Contriever \cite{gao2021unsupervised}, DPR \cite{karpukhin-etal-2020-dense}, and ANCE\cite{xiong2020approximate}. Following standard practice, we use dot product between the embedding vectors of a query/question and a candidate document as their similarity score \(R\).

\label{opinion-classfication}
(3) Opinion classification. We adopt Qwen2.5-Instruct-72B as the opinion classifier.
We selected high-performing LLMs that are widely recognized and adopted in the open-source community, based on the current GPU resources.
The detailed description is provided in Appendix \ref{exp-detail}

(4) Hyper-parameter settings. During the knowledge-guided attack process, we set the maximum editing distance $\epsilon$ to 0.2, the semantic similarity threshold $\lambda$ to 0.85, and the number of iterations $N$ to 5. For adversarial trigger generation, we use a beam size of 3, top-$k'$ sampling with $k'$ set to 10, a batch size of 32, a temperature of 1.0, a learning rate of 0.005, and a trigger sequence length of 10. In RAG\textsubscript{black} configuration, the number of retrieved documents $K$ is set to 3, and the LLM temperature is fixed at 1.0.

(5) Target documents. For the PROCON dataset, we rank documents based on their relevance to each topic-query set $Q$ and target stance $S_t$, then select the five least relevant documents as poisoning targets. For MS MARCO, we first retrieve the top-1000 passages ranked by relevance per topic, then identify the passage with the lowest average rank across the corresponding queries. This selection strategy ensures that evaluation is conducted under challenging conditions by focusing on minimally relevant passages.

(6) Environment. All the methods run on a server configured with Python 3.8 environment, four NVIDIA DGX A100 GPUs (80 GB each), and 1 TB of system memory.

\subsection{Baseline Settings}
To evaluate the effectiveness of our proposed method, we compare it against adversarial attack baselines tailored for black-box, topic-oriented RAG scenarios, with an emphasis on minimal modifications to the original documents. We exclude BadRAG\cite{xue2024badrag}, a backdoor RAG attack limited to white-box scenarios, as well as topic-IR-attack\cite{liu2023topic}, due to the lack of a complete implementation that prevents reliable reproduction.
For the baseline methods, we adapt them to meet the requirements of our task while preserving their core components. A brief overview of these baselines is provided below:

\textbf{PoisonedRAG.}
Zou et al.\cite{zou2024poisonedrag} propose an approach adaptable to both black-box and white-box settings. For our task, we employ its black-box strategy by inserting a randomly chosen query from the topic-queries set \( Q \) at the beginning of each document.

\textbf{PAT.}
This gradient-based adversarial retrieval attack \cite{liu2022order} uses a pairwise loss function to generate triggers that meet the fluency and coherence constraints. We adapt PAT to produce triggers \( T_{\text{pat}} \) for target documents within the topic-queries set, evaluating their effectiveness under black-box conditions.

\textbf{Collision.}
This method generates adversarial paragraphs (collisions) via gradient-based optimization to produce content semantically aligned with the target query. In a topic-queries context, we create collisions for the entire topic-queries set and examine their transfer performance on the retriever within black-box RAG.

These baseline methods provide benchmarks for comparing the efficacy of our approach in a black-box, topic-oriented RAG attack scenario.

\subsection{Evaluation Metrics}\label{sec:metrics}

For \textbf{RQ1}, which focuses on ranking manipulation, we report the change in top-3 target opinion proportion (\(\text{Top3}_{\text{att}} - \text{Top3}_{\text{ori}} = \text{top3-v}\)), Ranking Attack Success Rate (RASR), Boost Rank (BRank), and the percentage of target documents appearing in the Top-50 and Top-500. These metrics jointly reflect the effectiveness of rank promotion across queries.

\textbf{top3-v.} Computed by subtracting \(\text{Top3}_{\text{ori}}\) from \(\text{Top3}_{\text{att}}\), top3-v ranges from -1 to 1. A positive value signifies a successful increase of the target opinion in top-3 results, while a negative value indicates a detrimental effect.

\textbf{Ranking Attack Success Rate (RASR).} RASR captures how frequently target documents are successfully boosted in each query’s ranking. 

\textbf{Boost Rank (BRank).} BRank is the average rank improvement for all target documents under each query.

\textbf{Top-50, Top-500.} These metrics represent the percentage of target documents that move into specific ranking thresholds in the MS MARCO Dataset after manipulation. Higher percentages imply more effective promotion of target documents.

For \textbf{RQ2}, we employ Average Stance Variation (ASV) to quantify the extent to which the opinion manipulation influences LLM responses in a black-box RAG setting. To account for topic variability and inherent LLM diversity, we further introduce another evaluation metric Calibrated ASV ($\Delta$ASV).

\textbf{Average Stance Variation (ASV).}
ASV is the absolute difference between manipulated and original opinion scores (i.e., 0 = opposing, 1 = neutral, 2 = supporting). Higher values indicate stronger polarity shifts and manipulation impact.

\textbf{Calibrated ASV ($\Delta$ASV)}. To account for the inherent variability of controversial topics and the instability of LLM, we measure the baseline ASV in a clean state, denoted as ASV\textsubscript{clean} (calculated without adversarial manipulation). The adjusted score is then defined as  
\( \text{$\Delta$ASV} = \text{ASV} - \text{ASV\textsubscript{clean}} \). This adjustment ensures that $\Delta$ASV accurately captures the true impact of adversarial manipulation by removing the confounding effects of natural stance variation. It quantify the extent to which the polarity of the RAG-system outputs is affected by the manipulation. A positive $\Delta$ASV indicates a manipulation-induced polarity shift, with larger values reflecting stronger effectiveness.

\section{Experimental Results Analysis}

\begin{table}[!t]
  \centering
  \caption{White-box information retrieval (IR) attack performance (RASR, Top-50, and Top-500 are reported in \%) on MS MARCO dataset. \textit{Bold} indicates the best attack performance.}
  \resizebox{0.45\textwidth}{!}{
    \begin{tabular}{lcccc}
    \toprule
    Method & RASR & B-rank & Top-50 & Top-500 \\
    \midrule
    Collision & 93.19 & 245.19 & 0.46 & 40.56 \\
    PAT & 89.63 & 208.06 & 0.31 & 33.90 \\
    PoisonedRAG & 64.39 & 73.52 & 2.94 & 11.15 \\
    \textbf{Topic-FlipRAG} & \textbf{99.29} & \textbf{521.30} & \textbf{18.35} & \textbf{79.49} \\
    \bottomrule
    \end{tabular}
  \label{tab:ms_ir_attack_results}
  }
\end{table}

\begin{table*}[!t]
  \centering
  \caption{White-box and Black-box IR attack results (\%) on PROCON dataset. \textit{Bold} indicates the best attack performance.}
  \resizebox{0.7\textwidth}{!}{
    \begin{tabular}{cccccccccc}
    \toprule
    \multicolumn{2}{c}{\textbf{Setting}} & \multicolumn{2}{c}{\textbf{White-box}} & \multicolumn{6}{c}{\textbf{Black-box}} \\
    \cmidrule(r){1-2}\cmidrule(r){3-4}\cmidrule(r){5-10}
    \textbf{Target IR model} & \multirow{2}[4]{*}{\textbf{Target Stance}} & \multicolumn{2}{c}{\textbf{BERT}} & \multicolumn{2}{c}{\textbf{Contriever}} & \multicolumn{2}{c}{\textbf{ANCE}} & \multicolumn{2}{c}{\textbf{DPR}} \\
    \cmidrule(r){1-1}\cmidrule(r){3-4}\cmidrule(r){5-10} 
    \textbf{Method} &       & \textbf{RASR} & \textbf{top3-v} & \textbf{RASR} & \textbf{top3-v} & \textbf{RASR} & \textbf{top3-v} & \textbf{RASR} & \textbf{top3-v} \\
    \midrule
    \multirow{2}[2]{*}{Collision} & CON   & 48.44 & 23.23 & 21.27 & 8.38 & 18.79 & 10.41 & 16.87 & 5.97 \\
          & PRO   & 43.64 & 20.85 & 21.44 & 8.24 & 15.97 & 7.73 & 14.49 & 5.23 \\
    \midrule
    \multirow{2}[2]{*}{PAT} & CON   & 37.85 & 17.23 & 16.40 & 7.15 & 19.51 & 12.36 & 13.99 & 5.65 \\
          & PRO   & 31.99 & 13.35 & 13.07 & 4.30 & 13.44 & 9.14 & 8.61 & 2.26 \\
    \midrule
    \multirow{2}[2]{*}{PoisonedRAG} & CON   & 39.87 & 18.64 & 25.52 & 11.99 & 34.10 & 14.91 & 31.25 & 14.13 \\
          & PRO   & 32.75 & 14.04 & 22.04 & 8.71 & 25.37 & 9.81 & 21.26 & 8.31 \\
    \midrule
    \multirow{2}[2]{*}{Topic-FlipRAG} & CON   & \textbf{79.42} & \textbf{44.71} & \textbf{51.90} & \textbf{25.57} & \textbf{45.69} & \textbf{20.12} & \textbf{42.27} & \textbf{18.83} \\
          & PRO   & \textbf{75.01} & \textbf{37.04} & \textbf{49.20} & \textbf{21.55} & \textbf{42.12} & \textbf{20.57} & \textbf{41.84} & \textbf{16.41} \\
    \bottomrule
    \end{tabular}%
    }
  \label{tab:IR_attck_opinion}
\end{table*}

\begin{table*}[htbp]
  \centering
  \caption{Topic-Oriented RAG attack results on PROCON dataset, including ASV on different domains (Health \& Environment, Government \& Politics, Education, and Society \& Culture), average ASV, and $\Delta$ ASV. \textit{Bold} shows the best performances.}
    \resizebox{0.9\textwidth}{!}{
    \begin{tabular}{cccccccccccccc}
    \toprule
    \multirow{2}[4]{*}{\textbf{Method}} & \multirow{2}[4]{*}{\textbf{Target Stance}} & \multicolumn{5}{c}{\textbf{Contriever + Qwen2.5}} &       & \multicolumn{5}{c}{\textbf{Contriever + Llama3.1}} &  \\
\cmidrule(r){3-8}\cmidrule(r){9-14}          &       & \textbf{H\&E} & \textbf{G\&P} & \textbf{EDU} & \textbf{S\&C} & \textbf{avg. ASV} & \textbf{$\Delta$ ASV} & \textbf{H\&E} & \textbf{G\&P} & \textbf{EDU} & \textbf{S\&C} & \textbf{avg. ASV} & \textbf{$\Delta$ASV} \\
    \midrule
    \multirow{2}[2]{*}{Clean} & CON   & 0.21  & 0.27  & 0.24  & 0.23  & 0.24  & --    & 0.22  & 0.26  & 0.18  & 0.27  & 0.24  & -- \\
          & PRO   & 0.21  & 0.26  & 0.24  & 0.24  & 0.24  & --    & 0.23  & 0.27  & 0.21  & 0.27  & 0.25  & -- \\
    \midrule
    \multirow{2}[2]{*}{Collision} & CON   & 0.30  & 0.36  & 0.35  & 0.34  & 0.34  & 0.10  & 0.30  & 0.35  & 0.23  & 0.27  & 0.30  & 0.06  \\
          & PRO   & 0.27  & 0.40  & 0.30  & 0.28  & 0.32  & 0.08  & 0.29  & 0.37  & 0.24  & 0.28  & 0.31  & 0.06  \\
    \midrule
    \multirow{2}[2]{*}{PAT} & CON   & 0.30  & 0.27  & 0.28  & 0.27  & 0.28  & 0.04  & 0.26  & 0.35  & 0.25  & 0.32  & 0.30  & 0.06  \\
          & PRO   & 0.29  & 0.25  & 0.31  & 0.33  & 0.29  & 0.05  & 0.28  & 0.30  & 0.30  & 0.30  & 0.29  & 0.04  \\
    \midrule
    \multirow{2}[2]{*}{PoisonedRAG} & CON   & 0.30  & 0.35  & 0.26  & 0.38  & 0.33  & 0.09  & 0.34  & 0.39  & 0.31  & 0.46  & 0.38  & 0.14  \\
          & PRO   & 0.27  & 0.41  & 0.28  & 0.32  & 0.34  & 0.10  & 0.31  & 0.38  & 0.29  & 0.39  & 0.35  & 0.10  \\
    \midrule
    \multirow{2}[2]{*}{Topic-FlipRAG} & CON   & \textbf{0.51 } & \textbf{0.46 } & \textbf{0.48 } & \textbf{0.55 } & \textbf{0.49 } & \textbf{0.25 } & \textbf{0.54 } & \textbf{0.51 } & \textbf{0.30 } & \textbf{0.64 } & \textbf{0.50 } & \textbf{0.26 } \\
          & PRO   & \textbf{0.43 } & \textbf{0.55 } & \textbf{0.42 } & \textbf{0.51 } & \textbf{0.49 } & \textbf{0.25 } & \textbf{0.44 } & \textbf{0.58 } & \textbf{0.41 } & \textbf{0.44 } & \textbf{0.48 } & \textbf{0.24 } \\
    \bottomrule
    \end{tabular}%
    }
  \label{tab:rag-procon}%
\end{table*}%

\subsection{RQ1: Can Topic-FlipRAG effectively boost the rankings of target documents across-queries in the retriever within RAG?}

To explore RQ1, we first conducted white-box information retrieval (IR) manipulation experiments on  MSMARCO and PROCON datasets. The results of these experiments are presented in Table~\ref{tab:ms_ir_attack_results} and Table~\ref{tab:IR_attck_opinion}, respectively.

\textbf{MSMARCO Dataset.}
The results demonstrate that Topic-FlipRAG significantly outperforms all baselines across all evaluation metrics, even under challenging conditions where the MSMARCO dataset is configured with reduced intra-topic query similarity to simulate more extreme scenarios. As shown in Table~\ref{tab:ms_ir_attack_results}, it achieves a near-perfect RASR of 99.29\%, well above Poisoned-RAG (64.39\%), Collision (93.19\%), and PAT (89.63\%). It also records the highest B-rank (521.30) and strong top-ranked document coverage (18.35\% in top-50\%, 79.49\% in top-500\%), demonstrating its effectiveness in simultaneously manipulating rankings across multiple queries.

\textbf{PROCON Dataset.}
On the PROCON dataset (Table~\ref{tab:IR_attck_opinion}), the results also show the superiority of Topic-FlipRAG. \textbf{Under white-box settings}, for both target stances (CON and PRO), Topic-FlipRAG achieves the highest RASR (0.79 for CON, 0.75 for PRO) and the largest top3-v (0.45 for CON and 0.37 for PRO, respectively). In contrast, Poisoned-RAG, which uses a simple query+ strategy, shows considerably lower RASR (0.40 for CON and 0.33 for PRO, respectively), demonstrating the limitations of its approach for topic-oriented attacks. While the gradient-based methods, Collision and PAT, perform comparably to or even better than Poisoned-RAG in opinion manipulation, they still fall short of Topic-FlipRAG. 

To further evaluate our method’s practical effectiveness, we conducted \textbf{black-box IR manipulation} experiments on the PROCON dataset using three widely adopted dense retrievers, e.g., Contriever, ANCE, DPR. Owing to the inherent robustness of these retrievers, performance declined for all attack methods compared to the white-box setting. Nevertheless, Topic-FlipRAG displays the most consistent performance, retaining relatively high manipulation effectiveness across all models and metrics. For instance, under Contriever, Topic-FlipRAG achieves the highest RASR (0.52 for CON and 0.49 for PRO, respectively) and top3-v (0.26 for CON and 0.22 for PRO, respectively), significantly outperforming Poisoned-RAG, Collision, and PAT. A similar performance pattern is observed with ANCE and DPR, where Topic-FlipRAG consistently maintains a clear advantage.

Topic-FlipRAG's superior performance in transfer attack settings can be attributed to its integration of the LLM’s general semantic knowledge and NRM gradient guidance, along with a design specifically tailored for the topic-queries context. This combination enables the method to generate adversarial triggers that are both generalizable and contextually precise, allowing it to adapt effectively to varying retriever architectures without relying on the the parameters of the target IR model. In contrast, gradient-based methods like Collision and PAT depend heavily on gradient information from the target model, leading to substantial performance degradation in transfer attack scenarios. For instance, Collision achieves a maximum RASR of only 0.21 on Contriever for both CON and PRO stances, while PAT performs even worse, with RASR values of 0.16 (CON) and 0.13 (PRO).

Additionally, our findings reveal varying levels of vulnerability among dense retrievers. Contriever appears the most susceptible (exhibiting higher RASR and top3-v scores for all attackers), whereas DPR is notably more resilient (e.g., RASR for Topic-FlipRAG falls to 0.42). These observations underscore the importance of careful retriever selection when designing RAG systems to mitigate adversarial risks.

\subsection{RQ2: To what extent does Topic-FlipRAG affect the answers generated by the target RAG systems?}

Table~\ref{tab:rag-procon} shows the opinion manipulation results on RAG outputs, comparing two RAG systems: Contriever + Qwen2.5 and Contriever + Llama3.1. More results with other retrievers (ANCE and DPR) are reported in Table~\ref{asv_other_IR} (Appendix).

To evaluate the manipulation effectiveness across different topic domains, the PROCON dataset was divided into four subsets: Health \& Environment (H\&E), Government \& Politics (G\&P), Education (EDU), and Society \& Culture (S\&C). Each reported avg-ASV reflects the average manipulated stance polarity (PRO or CON) of the RAG outputs across all domains, while avg-$\Delta$ASV represents the manipulation effect, computed by subtracting the clean-state avg-ASV\textsubscript{clean} from the manipulated avg-ASV. Both ASV and $\Delta$ASV range from 0 to 2, where higher values indicate stronger manipulation effects.

The results demonstrate that \textbf{Topic-FlipRAG achieves significantly better manipulation across all domains compared to baseline methods}. Specifically, Topic-FlipRAG achieves an avg-ASV of 0.49 for Qwen2.5 and 0.48 for Llama3.1, outperforming all other methods. PoisonedRAG, the second most effective method, achieves avg-ASV values of 0.33 for Qwen2.5 and 0.35 for Llama3.1, while other baselines such as Collision and PAT remain within the range of 0.28 to 0.31. In terms of manipulation effect, measured by avg-$\Delta$ASV, Topic-FlipRAG also achieves the highest values: 0.26 for Qwen2.5 and 0.24 for Llama3.1.

Table~\ref{tab:rag-procon} and the appendix results further reveal notable differences between LLMs and retrievers regarding susceptibility to topic-oriented opinion manipulation. Overall, Llama3.1 exhibits higher avg-ASV and avg-$\Delta$ASV across all baseline methods compared to Qwen2.5, indicating that Llama3.1 is more vulnerable to adversarial documents promoted to the top-3 of the ranking. This suggests that Llama3.1 tends to generate responses that are more aligned with adversarial stances in the context. Similarly, among retrievers, contriever-based RAG systems achieve higher avg-$\Delta$ASV values compared to ANCE and DPR, as detailed in the appendix, aligning with RQ1 findings that contriever produces superior ranking distortion, particularly for top-3 results.

The manipulation effectiveness of Topic-FlipRAG varies across topic domains. For instance, with Llama3.1, S\&C questions are most prone to manipulation towards an opposing stance (avg-ASV = 0.55), whereas EDU questions are resistant to manipulation (avg-ASV = 0.41). It is more difficult to manipulate supporting stances in H\&E questions than opposing ones. Both Qwen2.5 and Llama3.1 achieve higher avg-ASV scores for supporting stance manipulation in G\&P, whereas EDU remains the most challenging domain.

In order to illustrate the manipulation results more vividly, we randomly selected the topic "Is Binge-Watching Good for You?" for a case study, as shown in Figure~\ref{case-study} (in Appendix). This topic comprises multiple sub-queries, enabling an examination of topic-level manipulation consistency. Before manipulation, responses span a mix of supportive, opposing, and neutral perspectives; after applying Topic-FlipRAG, the distribution shifts toward the target stance—becoming predominantly supportive under "Pro" (blue) or predominantly opposing under "Con" (red). For instance, in "Binge-watching: addictive or relaxing?", the original response presents a balanced view, while the manipulated versions adopt distinctly polarized stances—emphasizing relaxing effects under "Pro" and highlighting addictive harms under "Con." Similarly, for "Opinions on binge-watching trends?", the initially diverse responses become consistently biased toward the intended stance. These examples underscore Topic-FlipRAG's ability to impose coherent opinion shifts across multiple queries within a single topic.

\subsection{RQ3: Does topic-oriented opinion manipulation significantly impact users' perceptions of controversial topics?}

\begin{figure}[!t]
  \centering
  \includegraphics[width=0.43\textwidth]{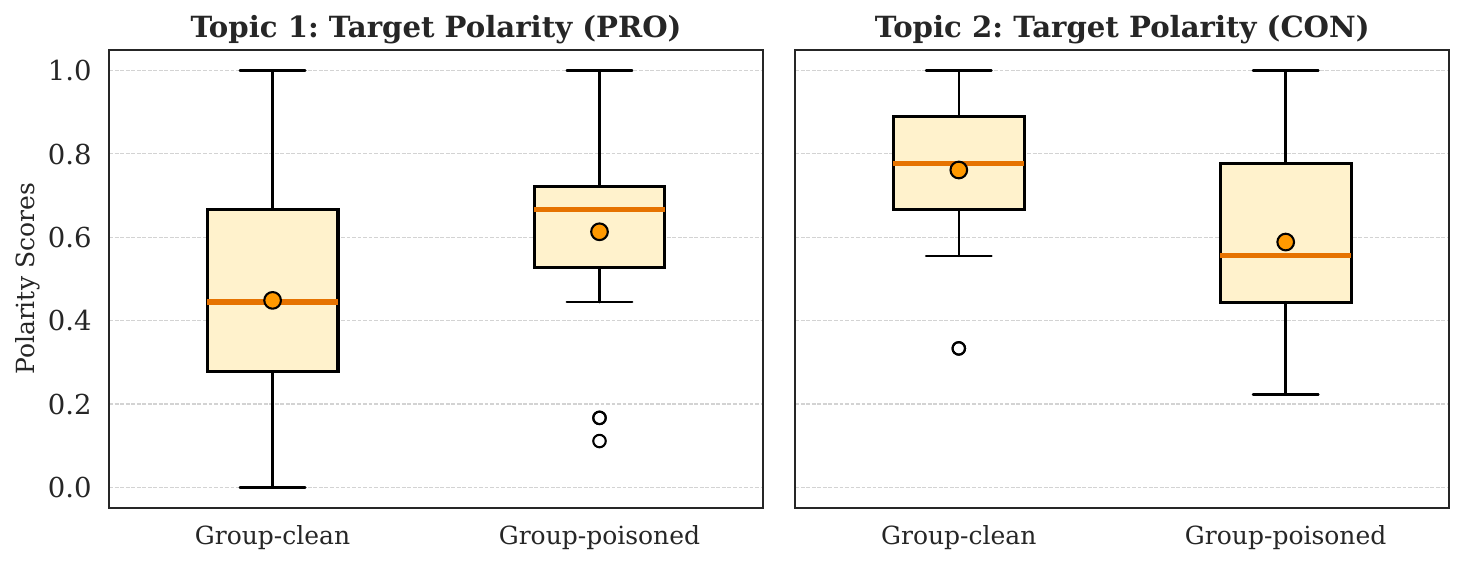}
  \caption{Empirical comparative experiments of adversarial opinion manipulation attacks on user cognition.}
  \label{user_experiment}
\end{figure}

To explore the effect of opinion manipulation on users’ perceptions of controversial topics, we conducted a user study involving 54 college students to assess how opinion manipulation influences perceptions of controversial topics. Participants were split into two equal groups (Group-clean and Group-poisoned, 27 members each), remaining unaware of the study’s true purpose and operating independently. A simple QA service based on RAG was implemented, and Topic-FlipRAG was employed to poison documents with specific stances on two designated controversial topics. For one topic, supporting documents were poisoned; for the other, opposing documents were poisoned. Each participant interacted with the QA system over three rounds per topic, mirroring real-world usage patterns, and subsequently rated their stance on a 7-point Likert scale (normalized to 0 for strong opposition, 1 for strong support). Full user experimental details are provided in the Appendix \ref{user-study-details}.

Figure~\ref{user_experiment} illustrates the significant shifts in user opinions resulting from Topic-FlipRAG. For Topic 1 (\textit{Should People Become Vegetarian?}), the manipulated stance was PRO. Group-clean exhibited a dispersed distribution with an average score of 0.45, reflecting diverse views on this controversial issue. By contrast, Group-poisoned recorded a notably higher mean score of 0.61, indicating successful manipulation toward pro-vegetarian sentiment. For Topic 2 (\textit{Should Humans Colonize Space?}), the stance was CON. Although Group-clean originally showed strong support (average score 0.76), Group-poisoned achieved a reduction to 0.59, effectively nudging opinions toward neutrality. These results underscore Topic-FlipRAG’s potency in shifting both neutral and polarized views, highlighting the potential risks of adversarial opinion manipulation in shaping public discourse.

\begin{table}[t]
  \centering
  \caption{Ablation study (top3-ori, top3-att, and top3-v are reported in \%) on the PROCON dataset. w/o denotes ``without''.}
  \resizebox{0.49\textwidth}{!}{
    \begin{tabular}{cccccccc}
    \toprule
    \multirow{2}[2]{*}{Method} & \multicolumn{3}{c}{IR} & \multicolumn{2}{c}{Llama3.1} & \multicolumn{2}{c}{Qwen2.5} \\
    \cmidrule(lr){2-4} \cmidrule(lr){5-6} \cmidrule(lr){7-8}
          & top3-ori & top3-att & top3-v & ASV & $\Delta$ASV & ASV & $\Delta$ASV \\
    \midrule
    Topic-FlipRAG & 42.83  & 70.24  & 27.41  & 0.64  & 0.37  & 0.55  & 0.32  \\
    w/o adv-trigger & 42.83  & 51.12  & 8.29  & 0.37  & 0.10  & 0.38  & 0.15  \\
    w/o know-attack & 42.83  & 59.40  & 16.57  & 0.40  & 0.12  & 0.41  & 0.18  \\
    clean & --    & --    & --    & 0.27  & --    & 0.23  & --    \\
    \bottomrule
    \end{tabular}%
  }
  \label{tab:ablation_study_component}
\end{table}

\subsection{Ablation Study}
\textbf{Effectiveness of Core Components of Topic-FlipRAG.} We investigated the individual contributions of the two main components of Topic-FlipRAG: know-attack and adversarial trigger generation (referred to as adv-trigger). As shown in Table~\ref{tab:ablation_study_component}, both components play significant roles in achieving the overall attack effectiveness, albeit with varying impacts on different aspects of the task.

Adv-trigger and know-attack complement each other by leveraging distinct strengths to enhance the overall effectiveness of the Topic-FlipRAG framework. Adv-trigger, through its gradient-based generation process, directly optimizes the retriever mechanism, resulting in significant ranking improvements for adversarially targeted documents. In contrast, know-attack, while comparatively less impactful on ranking optimization, utilizes general semantic knowledge to achieve precise and consistent polarity manipulation in LLM outputs, reinforcing the adversarial influence on generated responses. This synergy allows for more comprehensive and precise control over the RAG framework's behavior, thereby maximizing the overall efficacy of the attack strategy.

\textbf{Impact of Rewarding Function in Topic-FlipRAG.} The rewarding function in Topic-FlipRAG dynamically adjusts the augment factor \(t\) in the know-attack process, controlling the aggressiveness of edits in each iteration. By leveraging semantic similarity and edit distance filtering, the rewarding function ensures an adaptive improvement strategy, enhancing the overall performance.

As shown in Table~\ref{tab:reward-IR}, the rewarding function significantly impacts the effectiveness of know-attack in IR manipulation. Compared to fixed augment factor \(t\) values, the dynamic adjustment achieves higher attack breadth (evidenced by increased RASR) and depth (higher top3-v scores). Table~\ref{tab:reward-RAG} further highlights its influence on RAG opinion manipulation, where the dynamic approach yields higher ASV and \(\Delta\)ASV compared to fixed \(t\) settings. Notably, the rewarding function's influence on RAG output polarity manipulation surpasses its impact on IR manipulation. This can be attributed to its role in improving the quality of generated $\text{doc}_{know}$, where quality encompasses both enhanced ranking boosts (related to IR manipulation) and stronger polarity reinforcement. These findings underscore the rewarding function's critical contribution to optimizing the know-attack strategy for both ranking and opinion manipulation objectives.

\begin{table}[!t]
  \centering
  \caption{Comparative analyses of rewarding function and fixed augment factor $t$ on IR manipulation attacks (\%).}
  \resizebox{0.4\textwidth}{!}{
    \begin{tabular}{cccccc}
    \toprule

          & RASR & top3-ori & top3-att & top3-v \\
    \midrule
    know-attack & 22.25  & 42.83  & 52.12  & 8.29\\
    t=1     & 17.42  & 42.83  & 49.28  & 6.45 \\
    t=2     & 16.83  & 42.83  & 49.73  & 6.89 \\
    t=3     & 18.54  & 42.83  & 49.02  & 6.18 \\
    \bottomrule
    \end{tabular}%
  }
  \label{tab:reward-IR}
\end{table}

\begin{table}[t]
  \centering
  \caption{Comparative analyses of rewarding function and fixed augment factor $t$ on RAG opinion manipulation.}
  \resizebox{0.35\textwidth}{!}{
    \begin{tabular}{lcccc}
    \toprule
    Setting & Topic-FlipRAG & t=1 & t=2 & t=3 \\
    \midrule
      ASV& 0.64 & 0.58 & 0.56 & 0.59 \\
    $\Delta$ASV & 0.37 & 0.31 & 0.29 & 0.32 \\
    \bottomrule
    \end{tabular}%
  }
  \label{tab:reward-RAG}
\end{table}

\begin{figure}[!t]
  \centering
  \setlength{\abovecaptionskip}{4pt}   
  \setlength{\belowcaptionskip}{-2pt}   
  \includegraphics[width=0.43\textwidth]{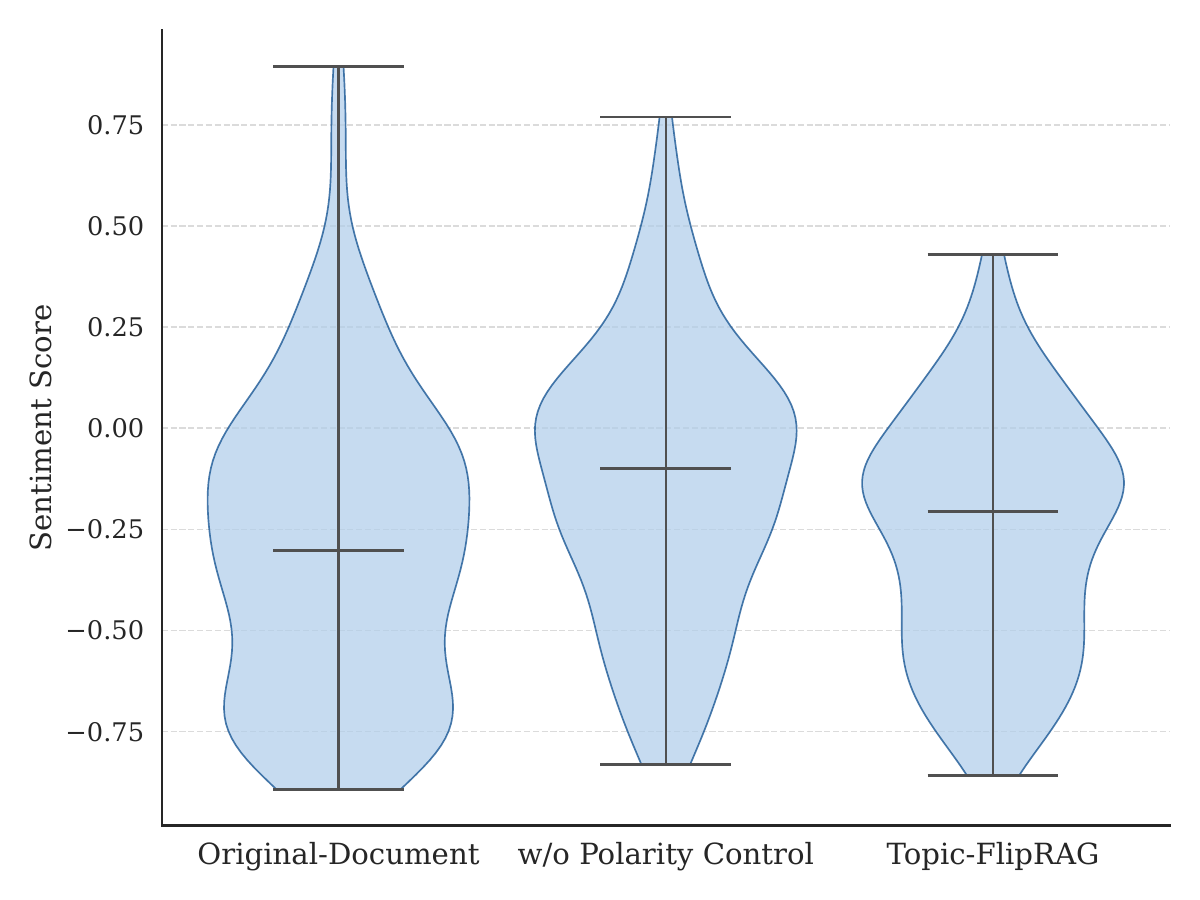}
  \caption{Impact of Polarity Control module on documents sentiment. w/o denotes ``without''. (target polarity:CON). }
  \label{sentiment-know}
\end{figure}

\textbf{Impact of Polarity Control in Topic-FlipRAG.} Figure~\ref{sentiment-know} illustrates the effect of the Polarity Control component on the sentiment of the $\text{doc}_{adv}$ when the target polarity is set to CON. Negative sentiment scores represent negative polarity, while positive scores indicate positive polarity. The results show that, without Polarity Control, the $\text{doc}_{adv}$ gravitates toward a neutral polarity. With Polarity Control enabled, Topic-FlipRAG effectively aligns $\text{doc}_{adv}$ with the desired target polarity CON).

The notable deviation in the \(\text{doc}_{adv}\) polarity without Polarity Control arises from the nature of know-attack: while it leverages the internal knowledge and reasoning capabilities of LLMs, it also inherits inherent polarity biases. Absent strict polarity enforcement during each editing step, the generated \(\text{doc}_{adv}\) tends to drift from the intended polarity. By enforcing the desired polarity at every phase of know-attack, the Polarity Control component supports a more consistent and effective manipulation of RAG polarity outcomes.

\begin{table}[!t]
  \centering
  \caption{Performance comparison of poisoning target on the PROCON dataset.(top3-ori, top3-att, top3-v are reported in \%.)}
  \resizebox{0.47\textwidth}{!}{
    \begin{tabular}{ccccccc}
    \toprule
    \multirow{2}[2]{*}{Poisoning Target} & \multicolumn{3}{c}{IR} & \multicolumn{2}{c}{Llama3.1} \\
    \cmidrule(lr){2-4} \cmidrule(lr){5-6}
          & top3-ori & top3-att & top3-v & ASV & $\Delta$ASV \\
    \midrule
    last5 & 42.83  & 70.24  & 27.41 & 0.64 & 0.37 \\
    top5  & 42.83  & 79.18  & 36.35 & 0.72 & 0.44 \\
    clean & --     & --     & --             & 0.27 & -- \\
    \bottomrule
    \end{tabular}%
  }
  \label{tab:sample_methods}
\end{table}

\begin{figure*}[!t]
  \centering
  \setlength{\abovecaptionskip}{2pt}   
  \setlength{\belowcaptionskip}{-4pt}   
  \includegraphics[width=0.97\textwidth]{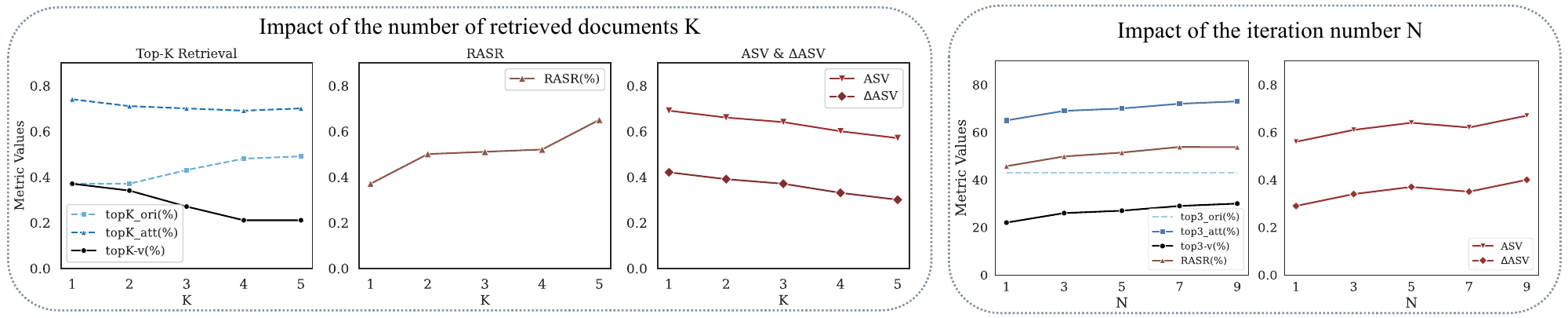}
  \caption{Impact of the number of retrieved documents $K$ (left part) and the iteration number $N$ (right part) on different performance metrics of Topic-FlipRAG on the PROCON dataset.}
  \label{parameter_K/N}
\end{figure*}

\textbf{Potential Effectiveness by Altering Poisoning Target.} Table~\ref{tab:sample_methods} illustrates the impact of different poisoned methods on IR and RAG manipulation results.In our prior experiments on the PROCON dataset, we intentionally selected the least relevant five documents (last5) as poisoning targets to assess Topic-FlipRAG's effectiveness under a challenging manipulation scenario. This design aimed to test the robustness of the attack under more constrained conditions. However, in real-world attack scenarios, an attacker might reasonably choose the topically most relevant documents (top5) as poisoned targets to achieve more effective opinion manipulation.

The experimental results show that poisoning the top5 documents yields improved manipulation performance, as evidenced by higher top3-v and ASV scores. Specifically, poisoning the top5 documents achieves a top3-v score of 0.37 and an ASV score of 0.71, both surpassing the performance observed with the last5 method. These results suggest that Topic-FlipRAG could pose even greater risks in real-world settings, where attackers can freely select highly relevant targets, highlighting the urgent need for robust defenses against such advanced adversarial strategies.

\subsection{Hyper-parameter Analysis}

We analyze four key hyper-parameters of our method: the editing distance $\epsilon$, the semantic similarity threshold $\lambda$, the iteration number $N$, and the number of retrieved documents $K$. Due to space limitations, the analyses of $\epsilon$ and $\lambda$ are demonstrated in detail in Appendix \ref{appendix:attack_param}. We do not discuss the hyper-parameters used in adversarial trigger generation, as they have already been systematically studied and validated in prior ranking attack research \cite{song2020adversarial,liu2022order}.

\textbf{Number of retrieved documents $K$.} This hyper-parameter controls how many candidate documents are retrieved for LLM generation. As $K$ varies, we replace the fixed top3-v metric with a dynamic topK-v to reflect stance manipulation across the top-$K$ candidates. RASR is also redefined accordingly: an attack is considered successful if the proportion of target-label documents increases within the top-$K$ results.

Our results (Figure \ref{parameter_K/N}) show that RASR increases with $K$, suggesting that when the retriever returns more documents, the chance of adversarial content appearing in the top-$K$ candidates rises, thus increasing the system’s vulnerability to manipulation. However, both topK-v and ASV exhibit a downward trend as $K$ increases. This indicates a dilution effect: while manipulated documents are more likely to be included, their influence on the final answer is weakened due to the expanded pool of candidates. To achieve a balanced evaluation that jointly accounts for both RASR and ASV, we set $K=3$ as the hyperparameter for our main experiments.

\textbf{Iteration number $N$.}
This hyper-parameter controls the total number of iterations in the knowledge-guided attack process, akin to training epochs in model optimization. As shown in Figure \ref{parameter_K/N}, increasing $N$ leads to improved performance across multiple evaluation metrics (e.g., ASV, RASR, top3\_att), indicating that more iterations enable the attack to better manipulate retrieved document stances and influence the final LLM output. For example, ASV improves from 0.56 at $N=1$ to 0.67 at $N=9$, while RASR follows a similar upward trend. 

However, higher $N$ also introduces greater computational cost. We observe that performance improves significantly from $N=1$ to $N=5$, but gains plateau between $N=5$ and $N=9$. To balance attack effectiveness and efficiency, we set $N=5$ as the default value in Topic-FlipRAG.

\section{Mitigation Analysis and Discussion}

To assess the robustness of existing defenses against topic-level opinion manipulation, we evaluate several strategies, including perplexity filtering, random masking, paraphrasing, and reranking. However, our empirical analysis shows that these methods are largely ineffective against Topic-FlipRAG, which leverages subtle, semantics-preserving perturbations that evade conventional detection and intervention. In light of these limitations, we further discuss potential defense directions, including utility-based filtering, TF-IDF anomaly detection, intra-top-$k$ similarity analysis, and certified provable mechanisms grounded in topic and fact consistency.

\subsection{Mitigation by Perplexity.}
Perplexity (PPL), a standard metric for text quality, is also used to detect adversarial attacks on LLMs and IR models\cite{jain2023baseline,gonen2022demystifying,liu2022order}. High perplexity indicates low-quality or suspicious text. We adopt PPL to identify malicious content.

\begin{figure*}[!t]
  \centering
  \includegraphics[width=0.95\textwidth]{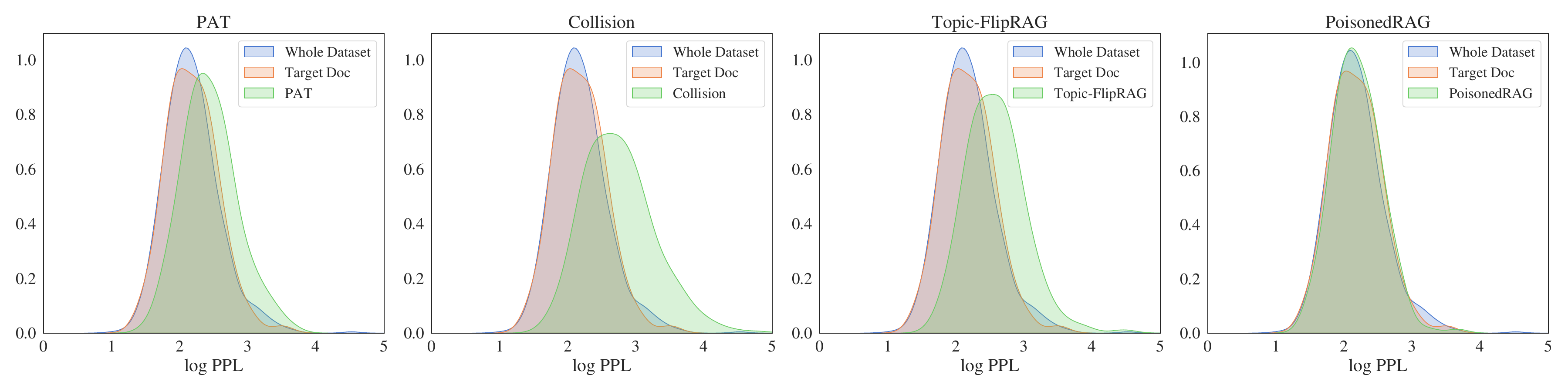}
  \caption{ Log perplexity (PPL) distributions on the PROCON dataset.}
  \label{ppl_defense}
\end{figure*}

Figure~\ref{ppl_defense} presents the GPT-2\cite{radford2019language}-evaluated log perplexity distributions of PROCON documents manipulated by various attack methods, highlighting the limitations of perplexity-based defenses. Due to substantial distributional overlap between clean and poisoned documents, setting effective thresholds proves challenging: strict thresholds yield high false positives, while loose ones miss attacks. Among the methods, Collision shows the most noticeable PPL shift, making it relatively detectable. In contrast, Topic-FlipRAG and PAT generate poisoned documents with distributions closely aligned to the original data, posing significant detection challenges. PoisonedRAG, lacking gradient-based optimization, induces negligible PPL changes, further underscoring the insufficiency of perplexity-based filtering for subtle manipulations.

\begin{figure}[!t]
  \centering
  \setlength{\abovecaptionskip}{4pt}   
  \setlength{\belowcaptionskip}{-2pt}
  \includegraphics[width=0.45\textwidth]{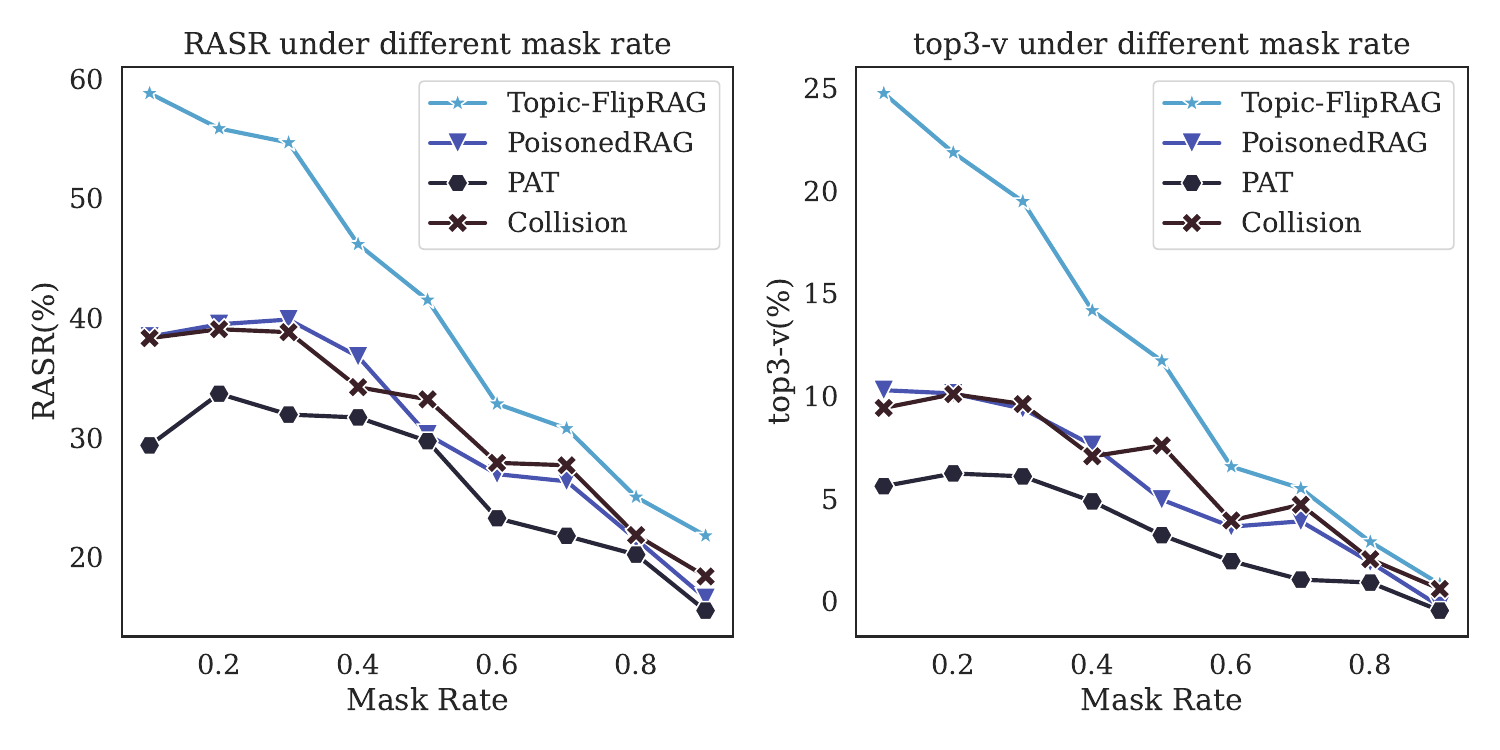}
  \caption{Attack performance of different random mask rate.}
  \label{random_mask}
\end{figure}

\subsection{Mitigation by Random Mask.} Random Mask is a robustness enhancement method that randomly masks a proportion of tokens with a placeholder token $m$ during embedding~\cite{zeng2023certified}. By averaging predictions over multiple masked input versions, it reduces sensitivity to small perturbations and enhances model stability. In our implementation, we averaged outputs over three masked copies of each input to assess performance under varying mask rates.

Figure~\ref{random_mask} presents experimental results on the PROCON-Dataset targeting the CON stance. When the mask rate is below 0.3, Topic-FlipRAG achieves notably higher manipulation metrics (RASR~>~55\%, top3-v~>~20\%) compared to other baselines, underscoring its robust manipulation capability. As the mask rate increases, the effectiveness of all methods declines; however, Topic-FlipRAG still maintains the highest attack success among the baselines. Nonetheless, excessively large mask rates also degrade normal RAG performance, revealing the trade-off between heightened defense and overall system functionality. Consequently, Random Mask alone is insufficient against sophisticated attacks like Topic-FlipRAG, indicating that additional measures are required to strengthen system robustness without undermining functionality.

\subsection{Mitigation by Paraphrasing}

Cheng et al.\cite{cheng2024trojanrag} and Zou et al. \cite{zou2024poisonedrag} investigate the application of paraphrasing defenses to evaluate the effectiveness of the RAG attacks they proposed. The underlying concept is to modify the query through rewriting, thereby increasing the difficulty for the attacker to successfully execute an attack on a specific query. The prompt used in our approach is provided in Appendix~\ref{mitigation_details}.

\begin{figure}[!t]
  \centering
  \includegraphics[width=0.47\textwidth]{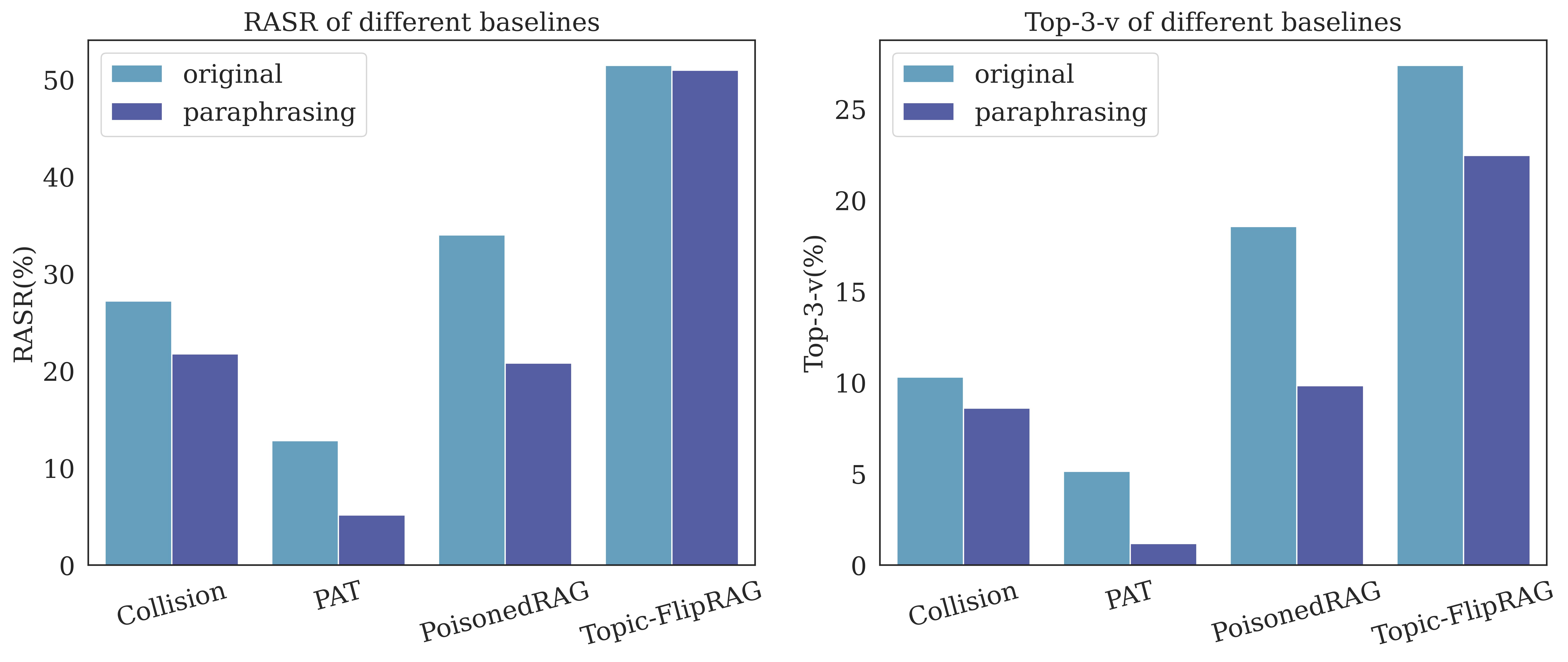}
  \caption{Comparative analysis of the performance metrics(RASR and top3-v) across different baselines before and after paraphrasing mitigation.}
  \label{paraphrasing}
\end{figure}

The attack effectiveness of Topic-FlipRAG and other baselines
against paraphrasing is presented in Table~\ref{paraphrasing}. Both the baselines and Topic-FlipRAG exhibit a decline in attack effectiveness when confronted with paraphrasing, yet Topic-FlipRAG remains largely unaffected: its RASR drops marginally from 51.47\% to 50.96\%, while top3-v decreases from 27.40\% to 22.46\%. These results suggest that paraphrasing is insufficient to defend against manipulations targeting the topical level. Topic-FlipRAG remains effective because it operates over entire topic-query sets rather than relying on specific query phrasings. Although paraphrasing alters surface wording, it typically preserves the original semantic focus, keeping the rewritten query within the same topic scope. This resilience underscores the difficulty of defending against topic-level RAG manipulations and highlights the need for more advanced, semantics-aware countermeasures.

\subsection{Mitigation by Reranking.}We investigate a reranking mechanism within the RAG framework to mitigate adversarial attacks. This approach employs a different IR model to reorder initial retrieval results, thereby reducing attack effectiveness. The defense rationale is twofold: (1) most attacks rely on approximating the target IR model~\cite{cho2024typos,liu2022order,liu2024multi}, and introducing a distinct reranker disrupts this dependency; (2) different IR models exhibit varying robustness, and reranking enhances overall resilience.

\begin{figure}[!t]
  \centering
  \includegraphics[width=0.49\textwidth]{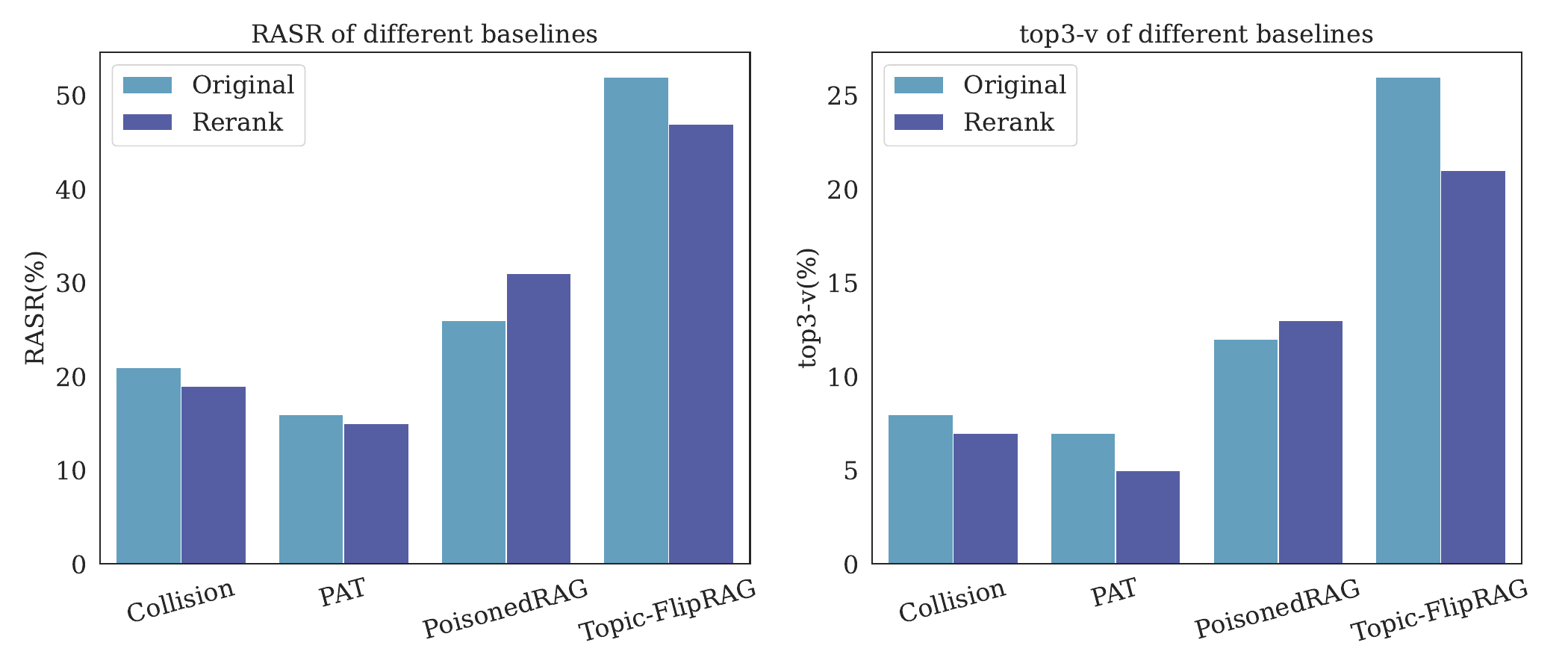}
  \caption{Comparative analysis of the performance metrics (RASR and top3-v) across different baselines before and after reranking mitigation.}
  \label{rerank}
\end{figure}

We tested this method on the PROCON-Dataset with the stance CON, using Contriever for retrieval and DPR for reranking. As illustrated in Figure~\ref{rerank}, most adversarial methods show a slight decline in manipulation effectiveness under reranking, except for PoisonedRAG, which is less sensitive due to its simpler query-plus strategy. In contrast, Collision, PAT, and Topic-FlipRAG depend on gradient information from open-sourced IR models, leading to varied transferability when attacking a different black-box IR pipeline. Despite the overall decrease in attack performance, Topic-FlipRAG maintains a relatively high success rate, demonstrating that reranking alone offers only partial defense against sophisticated adversarial techniques.

\subsection{Future Potential Defense Mechanisms}

We have integrated several widely adopted defense strategies from prior work on RAG-based and IR-based attacks, including filtering by perplexity, random masking, paraphrasing, and reranking-based optimization. While these techniques are commonly used, our empirical results demonstrate that they provide only limited robustness against the proposed attack method. To guide the development of more effective future defenses, we outline three promising directions informed by recent insights and attack characteristics:

\textbf{Filtering via Usefulness Features.}  
Recent RAG-related attack studies reveal that adversaries often enhance the \textit{relevance} of poisoned documents without improving their actual \textit{usefulness} in answering user queries. This decoupling creates opportunities for detection. For instance, RbFT~\cite{tu2025rbft} highlights the potential of usefulness-based filtering, such as leveraging LLM-based judgments or dedicated retrieval-utility models, to complement relevance-centric ranking systems. Incorporating usefulness estimation may significantly enhance filtering granularity in adversarial settings.

\textbf{TF-IDF Based Detection.}  
In topic-specific attack scenarios, adversaries frequently inject or substitute high-impact keywords to manipulate semantic relevance, leading to elevated TF-IDF scores. While such high TF-IDF values may also occur in naturally high-ranking documents, the distributional shift introduced by poisoned inputs, especially when paired with usefulness signals, could provide an effective basis for anomaly detection. This hybrid detection approach remains an open but promising research avenue.

\textbf{Intra-Top-\textit{k} Document Similarity.}  
Under benign retrieval conditions, the top-\textit{k} documents typically exhibit high semantic coherence, often reflected in strong pairwise similarity. In contrast, targeted poisoning tends to inject documents that diverge semantically from clean results, despite being top-ranked. This reduction in intra-group similarity offers a lightweight and model-agnostic signal that could flag manipulated retrieval outputs without additional supervision. Future work may explore leveraging this pattern for online detection in black-box RAG deployments.
\section{Conclusion}

This study highlights the critical vulnerability of RAG systems to sophisticated opinion manipulation attacks in real-world scenarios. By shifting focus from traditional factual or single-query attacks to topic-oriented adversarial manipulations, we reveal how attackers can systematically exploit the knowledge synthesis capabilities of LLMs to alter information perception and propagate biased narratives. The proposed Topic-FlipRAG framework demonstrates that combining adversarial ranking perturbations with LLMs' inherent reasoning abilities enables effective stance polarity shifts across interconnected queries, even in complex multi-perspective contexts. Experiments show that existing mitigation strategies inadequately address semantically sophisticated attacks. In the future, we will explore enhanced transparency protocols to maintain RAG systems as reliable information mediators.

\section*{Open Science}\label{open_science}
Our work complies with USENIX Security's open science mandate by prioritizing transparency and reproducibility. All attack implementations, including Topic-FlipRAG, evaluation scripts, and datasets, are hosted on GitHub\footnote{\url{https://github.com/LauJames/Topic-FlipRAG}} and Zenodo\footnote{\url{https://doi.org/10.5281/zenodo.15523435}}. The repository is licensed under the MIT License, allowing reuse while prohibiting malicious applications. We have provided comprehensive documentation, covering environment setup, hyper-parameters, and attack pipelines, to ensure independent verification. Sensitive data has been excluded from the repository.
Synthetic queries were generated using GPT-4o, with ethical constraints imposed to eliminate harmful or biased language. All components necessary to replicate our findings are publicly accessible. We are committed to maintaining the repository availability for at least three years and will update the repositories based on community feedback.

\section*{Ethics Considerations}
Our research investigates vulnerabilities in RAG systems through the lens of adversarial opinion manipulation. We rigorously evaluated ethical implications using principles from The Menlo Report 
(Beneficence, Respect for Persons, Justice, and Respect for Law) 
and the USENIX Security Ethics Guidelines. 

\textbf{Responsible Disclosure}. We understand the importance of engaging with model and framework providers, so we reported this vulnerability to the LangChain Security Team, hoping to collaborate on developing a mitigation module. However, due to limitations in defense measures, we have not yet gained recognition, and we are still striving to make effective progress. Additionally, the widespread use of hybrid open-source frameworks means that ownership boundaries are less clear compared to traditional software vulnerabilities. This ambiguity prompted us to explore a policy-oriented disclosure approach, aiming to encourage systemic preventative measures. Therefore, we used an accessible and efficient channel to initiate policy-level mitigation. We reported our preliminary findings and associated risks to our institution. This information, along with related content, was subsequently disclosed to Xinhua News Agency, not for public release. We want to emphasize that this initial disclosure was not intended to be exclusive or politically motivated. We are willing to inform other relevant platforms and international organizations.

\textbf{Data Privacy and User Consent.} Our experiments involved opinion-based datasets covering various topics. To protect individual privacy, we ensured that all datasets are anonymized to remove personally identifiable information (PII), thus safeguarding individual privacy. Moreover, before beginning the user experiments, participants were thoroughly informed about the study's background, purpose, and the measures in place to ensure fairness and privacy protection. After the experiment, we reiterated the true data distribution to participants to minimize any potential cognitive impact. Throughout the research, we conducted all experiments, particularly those involving adversarial attacks, in secure environments to prevent unintended consequences.

By addressing these ethical considerations, our research not only highlights vulnerabilities but also contributes to the development of robust solutions in safeguarding RAG models against exploitation.

\textbf{IRB Approval.}
This study was conducted under full Institutional Review Board (IRB) approval (Protocol No. \texttt{WHU-HSS-IRB2025010}. We confirm that all research activities strictly complied with the IRB-approved protocol. All research activities involving human participants received approval from our institution's Humanities and Social Sciences Ethics Committee. To obtain IRB approval, we submitted a comprehensive application package, including a research ethics justification report, informed consent forms, a detailed project protocol, and the CVs of both the principal investigator and all research participants. The application was subject to a rigorous, multi-stage review process involving sequential approval by the project leader, institutional supervisors, the ethics committee secretary, and the committee leadership. After successfully completing all review stages, our study was formally granted IRB approval. Specific procedures are detailed as follows:

\textit{Informed Consent:} All participants were required to sign an informed consent form prior to the experiment. The document clearly communicated the study’s objectives, procedures, estimated duration, potential psychological risks (e.g., exposure to controversial topics), compensation scheme (a cash reward of 50 RMB), and participants’ rights to withdraw at any time without penalty. Participants were also informed that their responses would be used solely for academic research.

\textit{Data Handling and Anonymization:} All responses and opinion scores were collected using anonymized participant IDs. No personally identifiable information (such as name, contact details, or personal ID) was stored. Data were encrypted and stored on secure servers accessible only to core research staff. Data will be retained for three years and then permanently deleted.
Throughout the study, strict anonymization procedures were enforced. Participant identifiers were randomly assigned, and all data records excluded any direct or indirect personal information. This ensured that individuals could not be re-identified during analysis or in any published findings.

\textit{User Privacy Protection:} The entire study was conducted in a controlled laboratory environment to minimize external data exposure. All data were used exclusively for academic purposes and were never shared with third parties. Participants were provided with post-experiment debriefing, including full disclosure of the study’s objectives and safeguards. Additionally, experimental group participants were shown control group outputs after completion to reduce potential cognitive bias introduced by adversarial content exposure.

In summary, this study not only adhered to rigorous ethical standards but also aimed to advance the development of trustworthy AI by systematically identifying vulnerabilities in RAG systems and promoting secure, transparent, and user-respecting model design.

\subsection*{Acknowledgement.} 
This work is supported by the National Science and Technology Major Project (2023ZD0121502) and National Natural Science Foundation of China (72404212).


\bibliographystyle{plain}

\bibliography{custom}
\FloatBarrier

\appendix

\section{Appendix}

\subsection{Additional  Experiment Details }\label{exp-detail}

(1) Opinion Classification.\label{opinion-judge} LLMs have been widely adopted as evaluators or "judges" in recent literature ~\cite{zheng2023judging}. Prior work has further validated the reliability of LLMs in sentiment classification~\cite{calderon2025alternative} and explicit stance detection~\cite{li2024advancing}.

In our experiment, we employ Qwen2.5-Instruct-72B as the opinion classifier. Owing to its large parameter scale and instruction tuning, the model demonstrates strong capability in identifying user stances embedded in natural language. To evaluate its reliability, we randomly sampled 500 RAG-generated responses from the PROCON dataset and recruited three cognitively unimpaired undergraduate students to independently annotate each instance. The Krippendorff's $\alpha$ among the human annotators was 0.8833, and the average pairwise Cohen's $\kappa$ between the LLM and the three human annotators was 0.7897.

We further aggregated the three annotations via majority vote to derive a human consensus label. The LLM's predictions achieved a $\kappa$ of 0.8110 and an accuracy of 0.8760 against this consensus, indicating strong agreement and supporting the use of LLMs as reliable stance evaluators in our setup.

\subsection{Additional Parameter Analysis}\label{appendix:attack_param}

\textbf{Similarity threshold $\lambda$.}
This parameter controls the minimum similarity between the poisoned document and the original one in the Know-Attack process. A lower value of $\lambda$ allows for greater perturbation and thus potentially more effective opinion manipulation. As shown in Table~\ref{tab:sim_threshold}, when $\lambda$ decreases from 0.90 to 0.80, we observe consistent improvements across multiple metrics: the top3-v score increases from 0.26 to 0.32, RASR improves from 0.51 to 0.58, and both ASV and $\Delta$ASV show significant increases (from 0.52 to 0.66 and from 0.25 to 0.39, respectively). These results indicate that permitting greater document alterations (i.e., lower $\lambda$) leads to more successful manipulation of both retrieval and generation stages.

\textbf{Edit distance $\varepsilon$.}
This parameter defines the allowed perturbation budget in the token space when generating poisoned documents. A larger edit distance permits more extensive token-level modifications, enabling the attacker to craft documents that are more semantically distant from the original. As shown in Table~\ref{tab:edit_distance}, increasing $\varepsilon$ from 0.10 to 0.30 consistently improves manipulation effectiveness. Specifically, the top3-v score rises from 0.25 to 0.30, and RASR improves from 0.47 to 0.56, indicating enhanced retrieval-stage manipulation. On the generation side, ASV increases from 0.57 to 0.65, while $\Delta$ASV grows from 0.30 to 0.38, suggesting that more aggressively perturbed documents induce greater shifts in LLM-generated opinions.

\begin{table}[t]
  \centering
  \caption{Performance under different similarity thresholds ($\lambda$) on the PROCON dataset. 
  Metrics top3-ori, top3-att, top3-v, and RASR are reported in \%.}
  \resizebox{0.47\textwidth}{!}{
    \begin{tabular}{ccccccc}
    \toprule
    \multirow{2}[2]{*}{similarity threshold $\lambda$} & \multicolumn{4}{c}{IR} & \multicolumn{2}{c}{Llama3.1} \\
    \cmidrule(lr){2-5} \cmidrule(lr){6-7}
          & top3-ori & top3-att & top3-v & RASR & ASV & $\Delta$ASV \\
    \midrule
    0.80  & 42.83  & 74.52  & 31.68  & 57.84 & 0.66 & 0.39 \\
    0.85  & 42.83  & 70.24  & 27.40  & 51.47 & 0.64 & 0.37 \\
    0.90  & 42.83  & 68.46  & 25.62  & 50.67 & 0.52 & 0.25 \\
    \bottomrule
    \end{tabular}%
  }
  \label{tab:sim_threshold}
\end{table}

\begin{table}[t]
  \centering
  \caption{Performance under different edit distances ($\epsilon$) on the PROCON dataset. 
  Metrics top3-ori, top3-att, top3-v, and RASR are reported in \%.}
  \resizebox{0.47\textwidth}{!}{
    \begin{tabular}{ccccccc}
    \toprule
    \multirow{2}[2]{*}{edit distance $\epsilon$} & \multicolumn{4}{c}{IR} & \multicolumn{2}{c}{Llama3.1} \\
    \cmidrule(lr){2-5} \cmidrule(lr){6-7}
          & top3-ori & top3-att & top3-v & RASR & ASV & $\Delta$ASV \\
    \midrule
    0.10  & 42.83  & 67.41  & 24.58  & 47.30 & 0.57 & 0.30 \\
    0.20  & 42.83  & 70.24  & 27.40  & 51.47 & 0.64 & 0.37 \\
    0.30  & 42.83  & 72.81  & 29.97  & 56.08 & 0.65 & 0.38 \\
    \bottomrule
    \end{tabular}%
  }
  \label{tab:edit_distance}
\end{table}

It is worth noting that a larger edit distance $\epsilon$ and a lower similarity threshold $\lambda$ both represent looser constraints on adversarial document modification. Both intuition and empirical results suggest that looser constraints lead to more effective manipulation. However, they may also make detection and mitigation easier, highlighting a trade-off between attack strength and stealth. Nevertheless, Topic-FlipRAG consistently outperforms the baseline across all evaluation metrics under various constraint settings. To balance attack effectiveness and applicability across different deployment scenarios, we adopt moderate default values of $\epsilon = 0.2$ and $\lambda = 0.85$ in our main experiments.

\subsection{User Study Details}\label{user-study-details}

To evaluate the effectiveness of adversarial opinion manipulation in real-world usage, we conducted a controlled user study involving 54 college students (aged 18--25; 29 males, 25 females). Participants were randomly assigned to two groups of equal size: the control group (Group-clean) and the experimental group (Group-poisoned) (27 participants each), and interacted independently with a QA system based on RAG system. Topic-FlipRAG was applied only for the experimental group to poison either supporting or opposing documents, enabling a between-subjects comparison of post-interaction stance shifts.

\textbf{Procedures.}
The experiment was fully computer-based and lasted approximately 30 minutes. Participants interacted with the system across two controversial topics. For each topic, participants first reported their initial stance using a 7-point Likert scale, which was later normalized to $[0,1]$ (where 0 = strong opposition, 1 = strong support). These pre-interaction scores were used to verify that there was no statistically significant difference in the baseline opinion distributions between the control group and the experimental group prior to any system exposure.

Afterward, the participants engaged in three rounds of interaction per topic. In each round, they were given a predefined query to input into the RAG-based QA system and were instructed to read the returned response. For the experimental group, the responses were manipulated using Topic-FlipRAG: one topic had its supporting documents poisoned, while the other had its opposing documents poisoned. The control group received unaltered responses. Following the interactions, participants again rated their opinions using the same 7-point Likert scale, enabling evaluation of manipulation effects.

\textbf{Participant Tasks.}
Participants were instructed to explore two controversial topics using the RAG-based QA system and to reflect on their personal stance. For each topic, they completed a two-stage opinion evaluation process. First, prior to any interaction with the system, they reported their initial stance using a 7-point Likert scale. Next, they engaged in three rounds of interaction with the RAG system, each involving a distinct predefined query. After reviewing all three responses, participants rated their post-interaction stance using the same 7-point scale.

To support the interaction process, we selected two topics from the PROCON-dataset and randomly sampled three sub-questions under each topic to serve as participant queries. Below are the exact sub-questions used in the study.

\begin{description}[leftmargin=0.5cm, labelindent=0cm]
  \item[\textbf{Topic 1:}] \textit{Should People Become Vegetarian?}\\
  Q1: Is vegetarianism compassionate? \\
  Q2: Is vegetarianism influential? \\
  Q3: Should teenagers be vegetarian?
  \item[\textbf{Topic 2:}] \textit{Should Humans Colonize Space?}\\
  Q1: Should humans pursue extraterrestrial colonization? \\
  Q2: Are humans ready for space colonization? \\
  Q3: What are your thoughts on space colonization?
\end{description}

The following evaluation questions were used to formally measure participants' stance on each topic, both before and after their interaction with the QA system:

\begin{description}[leftmargin=0.5cm, labelindent=0cm]
  \item[\textbf{Topic 1:}] \textit{Should People Become Vegetarian?}\\
  \textbf{Pre-interaction:} \textit{"For you personally, do you think vegetarianism is beneficial or meaningless?"} \\
  \textbf{Post-interaction:} \textit{"After reading the content on this topic, do you think vegetarianism is meaningful? Please rate the degree of necessity."}

  \item[\textbf{Topic 2:}] \textit{Should Humans Colonize Space?}\\
  \textbf{Pre-interaction:} \textit{"For you personally, do you think humanity's current investment in interstellar exploration is more beneficial or more harmful?"} \\
  \textbf{Post-interaction:} \textit{"After learning about this topic, assuming you are a member of the senate, would you support or oppose large-scale societal investment in space expansion?"}
\end{description}

\textbf{Pre/Post-Inform Protocol.}
To avoid priming effects, participants were initially informed that the study aimed to "evaluate user cognitive feedback on AI QA systems," with no mention of adversarial attacks or opinion manipulation. This omission was deliberate to preserve the naturalness of user responses. Following the experiment, a full debriefing was provided in accordance with IRB guidelines. Participants in the experimental group were informed of the adversarial modifications applied to system outputs, and all participants were given the option to review the clean system responses used for the control group. Contact information was provided for follow-up inquiries.

\begin{table}[h]
\caption{Pre-Interaction Opinion Score Comparison Between Control and Experimental Groups Across Topics}
\centering
\resizebox{0.43\textwidth}{!}{

\begin{tabular}{llcc}
\toprule
\textbf{Topic} & \textbf{Group} & \textbf{Mean ± SD} & \textbf{\textit{p}-value} \\
\midrule
\multirow{2}{*}{Topic 1} 
  & Control & 0.4671 ± 0.2153 & \multirow{2}{*}{0.6839} \\
  & Experimental & 0.4424 ± 0.2278 & \\
\midrule
\multirow{2}{*}{Topic 2} 
  & Control & 0.6996 ± 0.1690 & \multirow{2}{*}{0.9370} \\
  & Experimental & 0.7037 ± 0.2084 & \\
\bottomrule
\end{tabular}}
\label{tab:group_stats}
\end{table}

\textbf{Statistical Analysis.}
Before interacting with the QA system, participants in both groups rated their initial stance on each topic. As shown in Table~\ref{tab:group_stats}, the means and variances of pre-interaction scores were highly similar across the two groups, indicating that participants began with comparable baseline opinions. To statistically assess baseline equivalence, we tested the null hypothesis $\bm{H_0}$: \textbf{\textit{the control and experimental groups do not exhibit significant differences in their initial scores}}. For both topics, the resulting \textit{p}-values were substantially greater than 0.05, indicating that $\bm{H_0}$ cannot be rejected and supporting the conclusion that no significant differences exist between groups prior to the intervention. These results demonstrate the effectiveness of random assignment and rule out initial group bias.


To assess the impact of adversarial document poisoning, we conducted independent two-sample \textit{t}-tests on post-interaction stance scores. Significant opinion shifts were observed in the experimental group compared to the control group: for Topic 1 (Vegetarianism), $p = 0.0454 < 0.05$; and for Topic 2 (Space Colonization), $p = 0.0017 < 0.005$.

\subsection{Additional Mitigation Details}\label{mitigation_details}

Our paragraphing approach uses the following prompt:

\begin{tcolorbox}
    \textbf{Task:} Rewrite the following query while preserving its original meaning. Aim to modify as many words and expressions as possible, while ensuring the intent remains intact. \\
    \textbf{Original Question:} \{question\}
\end{tcolorbox}

\begin{table*}[!t]
  \centering
  \caption{Average stance variation (ASV) and $\Delta$ASV for different attack methods across four retriever-generator combinations. \textbf{Bold} indicates best performance in each setting.}
  \resizebox{0.85\textwidth}{!}{
  \begin{tabular}{cccccccccc}
    \toprule
    \multirow{2}[4]{*}{\textbf{Method}} & \multirow{2}[4]{*}{\textbf{Target Stance}}

    & \multicolumn{2}{c}{\textbf{ANCE + Qwen2.5}} 
    & \multicolumn{2}{c}{\textbf{ANCE + Llama3.1}} 
    & \multicolumn{2}{c}{\textbf{DPR + Qwen2.5}} 
    & \multicolumn{2}{c}{\textbf{DPR + Llama3.1}} \\
    \cmidrule(r){3-4} \cmidrule(r){5-6} \cmidrule(r){7-8} \cmidrule(r){9-10}
    & & \textbf{avg. ASV} & \textbf{$\Delta$ASV} & \textbf{avg. ASV} & \textbf{$\Delta$ASV}
    & \textbf{avg. ASV} & \textbf{$\Delta$ASV} & \textbf{avg. ASV} & \textbf{$\Delta$ASV} \\
    \midrule
    \multirow{2}{*}{Clean} 
    & CON & 0.21 & -- & 0.24 & -- & 0.25 & -- & 0.23 & -- \\
    & PRO & 0.21 & -- & 0.24 & -- & 0.24 & -- & 0.24 & -- \\
    \midrule

    \multirow{2}{*}{Collision} 
    & CON & 0.28 & 0.06 & 0.32 & 0.08 & 0.27 & 0.02 & 0.33 & 0.10 \\
    & PRO & 0.29 & 0.08 & 0.34 & 0.10 & 0.31 & 0.07 & 0.35 & 0.12 \\
    \midrule
    \multirow{2}{*}{PAT} 
    & CON & 0.25 & 0.04 & 0.29 & 0.05 & 0.26 & 0.05 & 0.28 & 0.05 \\
    & PRO & 0.25 & 0.05 & 0.31 & 0.07 & 0.25 & 0.01 & 0.27 & 0.04 \\
    \midrule
    \multirow{2}{*}{PoisonedRAG} 
    & CON & 0.35 & 0.13 & 0.36 & 0.13 & 0.37 & 0.12 & 0.43 & 0.20 \\
    & PRO & 0.32 & 0.12 & 0.33 & 0.09 & 0.32 & 0.08 & 0.37 & 0.14 \\
    \midrule
    \multirow{2}{*}{Topic-FlipRAG} 
    & CON & \textbf{0.42} & \textbf{0.21} & \textbf{0.46} & \textbf{0.23} & \textbf{0.44} & \textbf{0.19} & \textbf{0.46} & \textbf{0.23} \\
     & PRO & \textbf{0.39} & \textbf{0.18} & \textbf{0.47} & \textbf{0.23} & \textbf{0.46} & \textbf{0.21} & \textbf{0.44} & \textbf{0.21} \\
    \bottomrule
  \end{tabular}
  
  \label{asv_other_IR}
  }
\end{table*}

\begin{figure*}[!t]
  \centering
  \includegraphics[width=0.95\textwidth]{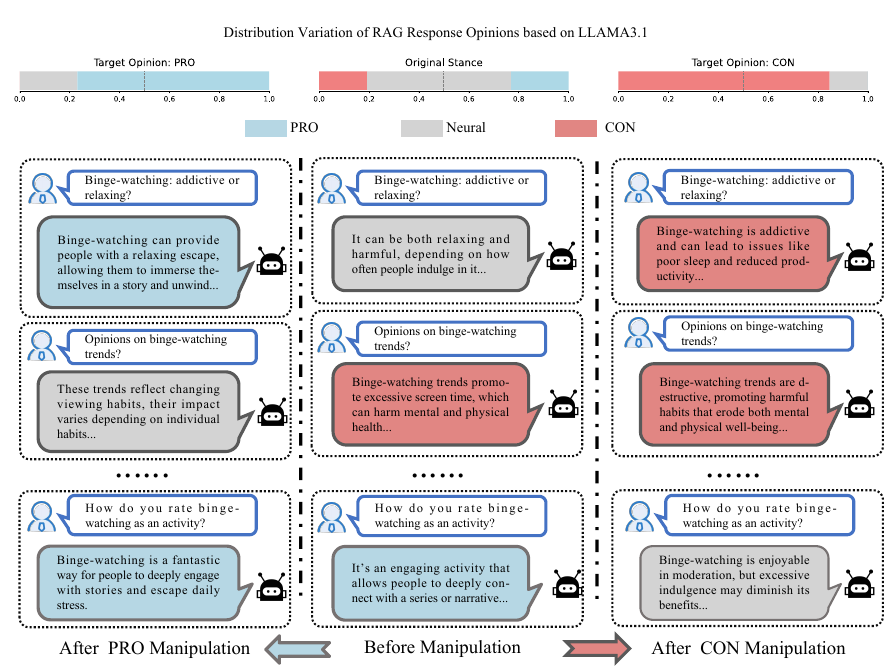}
  \caption{Case study of Topic-oriented Opinion Manipulation On Topic:"Is Binge-Watching Good for You?".}
  \label{case-study}
\end{figure*}

\subsection{Additional Manipulation Analyses}

To further explore RQ2: \textit{To what extent does Topic-FlipRAG affect the answers generated by the target RAG systems?}, we conducted RAG-based opinion manipulation experiments on the PROCON dataset using ANCE and DPR as retrievers. Detailed experimental outcomes are provided in Table~\ref{asv_other_IR}. Figure~\ref{case-study} represents the case-study based on Topic: "Is Binge-Watching Good for You?"

\end{document}